\documentclass{article}
\usepackage[preprint]{neurips_2024}

\usepackage[utf8]{inputenc} % allow utf-8 input
\usepackage[T1]{fontenc}    % use 8-bit T1 fonts
\usepackage{hyperref}       % hyperlinks
\usepackage{url}            % simple URL typesetting
\usepackage{booktabs}       % professional-quality tables
\usepackage{amsfonts}       % blackboard math symbols
\usepackage{nicefrac}       % compact symbols for 1/2, etc.
\usepackage{microtype}      % microtypography
\usepackage{xcolor}         % colors

\usepackage{bibunits}
\usepackage{times}
\usepackage{helvet}
\usepackage{courier}
\usepackage{graphicx}
\usepackage{caption}
\usepackage{algorithm}
\usepackage{algorithmic}
\usepackage{newfloat}
\usepackage{listings}
\DeclareCaptionStyle{ruled}{labelfont=normalfont,labelsep=colon,strut=off}
\floatstyle{ruled}
\newfloat{listing}{tb}{lst}{}
\floatname{listing}{Listing}

\usepackage{amsmath}
\usepackage{amsfonts}
\usepackage{amssymb}
\usepackage{amsthm}
\usepackage{braket}
\usepackage[capitalize]{cleveref}
\usepackage{tikz}
\usepackage{mathtools}
\usepackage{ifthen}

\newcommand{\anonymousOrNot}[2]{%
  \ifthenelse{\boolean{@submission}}%
    {#1}% If in submission mode
    {#2}% If in preprint or final mode
}

\newcommand{\citeCompanionPaper}[1][NULL]{%
  \ifthenelse{\equal{#1}{NULL}}%
    {\ifthenelse{\boolean{@submission}}%
      {\cite{anonymous2024classifim-physics}}% If in submission mode
      {\cite{kasatkin2024classifim-physics}}}% If in preprint or final mode
    {\ifthenelse{\boolean{@submission}}%
      {\cite[#1]{anonymous2024classifim-physics}}% If in submission mode
      {\cite[#1]{kasatkin2024classifim-physics}}}% If in preprint or final mode
}

\newcommand{\citeImplementation}{%
  \ifthenelse{\boolean{@submission}}%
    {\cite{anonymous2024classifim-code}}% If in submission mode
    {\cite{kasatkin2024classifim-code}}% If in preprint or final mode
}

\newtheorem{theorem}{Theorem}

\DeclareMathOperator{\distMSE}{distMSE}
\DeclareMathOperator{\distMSEPS}{\distMSE_{\textnormal{PS}}}
\DeclareMathOperator{\distRE}{distRE}
\DeclareMathOperator{\dist}{dist}
\DeclareMathOperator{\mean}{mean}

\DeclareMathOperator{\Tr}{Tr}
\newcommand{\floor}[1]{\left\lfloor #1 \right\rfloor}
\newcommand{\ceil}[1]{\left\lceil #1 \right\rceil}
\newcommand{\myvec}[1]{{\boldsymbol{#1}}}
\newcommand{\vlambda}{\myvec{\lambda}}
\newcommand{\dlambda}{\delta\lambda}
\newcommand{\vdlambda}{\myvec{\dlambda}}
\newcommand{\vecv}{\myvec{v}}
\newcommand{\vecl}{\myvec{l}}
\newcommand{\veczero}{\myvec{0}}
\newcommand{\vectheta}{\myvec{\theta}}
\newcommand{\distSL}{\dist_{\textnormal{SL}}}
\newcommand{\abs}[1]{\left|#1\right|}
\newcommand{\norm}[1]{\left\|#1\right\|}

\newcommand{\mcM}{\mathcal{M}}
\newcommand{\mcS}{\mathcal{S}}
\newcommand{\mcD}{\mathcal{D}}
\newcommand{\mcDtrain}{\mathcal{D}_{\textnormal{train}}}
\newcommand{\mcDtest}{\mathcal{D}_{\textnormal{test}}}
\newcommand{\mcDtrainBC}{\mathcal{D}_{\textnormal{train,BC}}}
\newcommand{\mcDBC}{\mathcal{D}_{\textnormal{BC}}}

\title{ClassiFIM: An Unsupervised Method To Detect Phase Transitions}

\author{
Victor Kasatkin \\
USC Viterbi \\
\texttt{postdoc@worldofml.com} \\
\And
Evgeny Mozgunov \\
USC Viterbi ISI \\
\And
Nicholas Ezzell \\
USC \\
\AND
Utkarsh Mishra \\
University of Delhi \\
\And
Itay Hen \\
USC \\
\And
Daniel Lidar \\
USC}

\begin{document}

\begin{bibunit}[abbrvnat]

\maketitle

\begin{abstract}
Estimation of the Fisher Information Metric (FIM-estimation) is an important task that arises in unsupervised learning of phase transitions, a problem proposed by physicists. This work completes the definition of the task by defining rigorous evaluation metrics $\distMSE$, $\distMSEPS$, and $\distRE$ and introduces ClassiFIM, a novel machine learning method designed to solve the FIM-estimation task. Unlike existing methods for unsupervised learning of phase transitions, ClassiFIM directly estimates a well-defined quantity (the FIM), allowing it to be rigorously compared to any present and future other methods that estimate the same. ClassiFIM transforms a dataset for the FIM-estimation task into a dataset for an auxiliary binary classification task and involves selecting and training a model for the latter. We prove that the output of ClassiFIM approaches the exact FIM in the limit of infinite dataset size and under certain regularity conditions. We implement ClassiFIM on multiple datasets, including datasets describing classical and quantum phase transitions, and find that it achieves a good ground truth approximation with modest computational resources. Furthermore, we independently implement two alternative state-of-the-art methods for unsupervised estimation of phase transition locations on the same datasets and find that ClassiFIM predicts such locations at least as well as these other methods. To emphasize the generality of our method, we also propose and generate the MNIST-CNN dataset, which consists of the output of CNNs trained on MNIST for different hyperparameter choices. Using ClassiFIM on this dataset suggests there is a phase transition in the distribution of image-prediction pairs for CNNs trained on MNIST, demonstrating the broad scope of FIM-estimation beyond physics.
\end{abstract}

\section{Introduction}
The task of estimating the Fisher Information Metric (FIM) and its relationship with unsupervised detection of both classical and quantum phase transitions is of great interest to physicists seeking to discover new phases of matter. A companion paper \citeCompanionPaper{} introduces this task and describes the generation and analysis of corresponding physics datasets. FIM estimation with a 1-dimensional parameter space is related to the change point detection task (defined, e.g., in \cite{truong2020selective}) when data points can be interpreted as samples from a probability distribution.
We note that the FIM was also used in machine learning, albeit in a different context \citeCompanionPaper[App. G]{}: see, e.g., \cite{karakida2019universal,martens2020new,kirkpatrick2017overcoming,amari1998natural}.
Here, we introduce ClassiFIM, a method for solving the FIM-estimation task, evaluate it on a number of datasets, and compare it to existing state-of-the-art methods {\NoHyper{\cite{vanNieuwenburg2016LearningPT, Huang:22}}} (even though they are designed to solve different tasks, also related to unsupervised learning of phase transitions) on the same datasets.

This work and the companion paper \citeCompanionPaper{} are complementary, written for different audiences, and can be read independently. This work is aimed primarily at computer scientists, while the companion paper is aimed primarily at physicists.

\section{Background on Fisher information metric}
\label{sec:fim}
In order to formally define our task, we must first define the
Fisher information metric (FIM), which is a formal way to measure the square of the rate of
change of a probability distribution. Consider a statistical manifold (SM), i.e.
a collection of distributions $P_{\vlambda}(\bullet)$,
depending on a vector of continuous parameters $\vlambda=(\lambda_1,\cdots,\lambda_N)$,
satisfying certain regularity conditions \cite{Schervish:1995aa}.
The score function for the $\mu$-th parameter $\lambda_{\mu}$ is the derivative
of the log-likelihood and is given by
$s_{\mu}(x;\vlambda) = \partial_{\lambda_{\mu}} \ln P_{\vlambda}(x)$.
The Fisher Information Metric (FIM) is the variance of the score function or,
equivalently (since by normalization
$\mathbb{E}[s_{\mu}(X;\vlambda)]=\int dx\, P_{\vlambda}(x)s_{\mu}(x;\vlambda)=0$),
the expectation value of the product of pairs of score functions:
\begin{equation}
  \label{eq:fim}
  g_{\mu\nu}(\vlambda) = \mathbb{E}[s_{\mu}(X;\vlambda)s_{\nu}(X;\vlambda)] .
\end{equation}
Chencov's theorem implies that $g_{\mu\nu}(\vlambda)$ is (up to rescaling)
the unique Riemannian metric on the space of probability distributions
satisfying a monotonicity condition and invariant under taking sufficient
statistics \cite{chencov1972statistical,amari2016information}.
That is, the FIM is the natural way to measure (the square of) the speed with which the
probability distribution changes with respect to the parameters $\vlambda$.
In fact, the FIM (up to a scaling factor) can be written as the first
non-trivial term in the Taylor series expansion in the parameter differences
$\vdlambda$ of multiple divergence measures
$D(P_{\vlambda}||P_{\vlambda+\vdlambda})$, including the Kullback-Leibler
divergence and the Jensen-Shannon divergence.

The FIM is a (non-negative definite) Riemannian metric on the parameter space
$\mcM$, i.e., it can be naturally used to measure lengths of curves in the
parameter space.

The above coordinate description can be generalized to $\vlambda$ belonging to
a smooth manifold $\mcM$, resulting in a symmetric 2-form $g$, which is the FIM,
and the ClassiFIM method would still be applicable.
In the remainder of this work, however, we, for simplicity,
focus on the case $\mcM = [0, 1]^N$ for $N=1,2$ and the case where the dataset
contains samples with $\vlambda_i \in \mcM' \subset \mcM$, where $\mcM'$
is a regular square grid in $\mcM$.

More details on the FIM can be found in \citeCompanionPaper[Sec. II]{} and in \cref{as:fim}.

\section{The FIM-estimation task}
\label{ss:fim-estimation-task}
The FIM-estimation task was proposed in \citeCompanionPaper[Sec III.A]{}.
Here we remind the reader of the task and propose the evaluation metrics,
thus completing the formal definition of the task.

Throughout this work, the input dataset is associated with an SM $(\mcM, P)$
and has the form
$\mcDtrain = \{(\vlambda_i, x_i)\}_{i=1}^{\abs{\mcDtrain}}$,
where each $x_i$ is a sample from the probability distribution $P_{\vlambda_i}(\bullet)$.
In addition to the dataset, some information about the nature and the data format
of the samples $x_i$ might be given (see \cref{as:ds-description} for an example).
The task is to provide an estimate $\hat{g}_{\mu\nu}(\vlambda)$ of the FIM
for all $\vlambda \in \mcM$.

When the ground truth FIM is available, one can measure the quality
of such an estimate using the metrics $\distMSE$, $\distMSEPS$, and $\distRE$
introduced below. We provide more details, including the precise definitions
of each of the metrics in \cref{as:benchmarking}.
The key property of the evaluation metrics is that they estimate how well the
estimated FIM reproduces the distances between points in the parameter space.
In order to compute each of the metrics,
one needs to compute the lengths of straight line intervals between all
$\abs{\mcM'}(\abs{\mcM'}-1)/2$ pairs of points in $\mcM'$ according to both
the ground truth FIM $g$ and the estimated FIM $\hat{g}$.
Then, $\distRE$ is (up to minor technical details) the probability that the
estimated FIM ranks two such lengths incorrectly. $\distMSE$ is the mean squared error
between the lengths according to the estimated and the ground truth FIMs.
$\distMSEPS$ is a variant of $\distMSE$ that is invariant under rescaling
of the estimated FIM: it is equal to the minimum of $\distMSE$ of a scaled
prediction taken over all positive (constant) scaling factors.

When evaluating the performance of a method, one might use $\distMSE$
when well-calibrated (correctly scaled) predictions are needed
and $\distMSEPS$ or $\distRE$ when the scale of the predictions is less important.
$\distRE$ is always between 0 and 1 ($100\%$) and is independent
of the scale of the predictions or the ground truth.

\section{The ClassiFIM method}
\label{sec:classifim}

We propose ClassiFIM as an \emph{unsupervised} ML method to solve FIM-estimation tasks.

\subsection{Idea behind ClassiFIM}

Consider a simple case where the probability distribution $P_{\lambda}$ is parameterized by a single
parameter $\lambda$ and we wish to estimate the FIM at $\lambda = 0$. The idea is to
pick a small $\varepsilon$ and randomly select a sign $\pm$ (with probability $0.5$ each),
then sample $x$ from $P_{\lambda=\pm\varepsilon}$ and hand both $\varepsilon$
and $x$ to a trained binary classification model (i.e., a binary classifier (BC)) trying to predict what the sign was.
If the distributions $P_{\lambda=\pm\varepsilon}$ are identical, the FIM is likely
to be small and a well-trained model trained to minimize the cross-entropy
should return log odds close to $0$. On the other hand, if these distributions
are very different, the model should be confident of what the sign was and the log odds
should be large by absolute value. As the careful derivation in \cref{th:bitchifc} below will
show (in a more general setting),
the mean of square of the log odds divided by $4\varepsilon^2$ indeed converges
to the FIM at $\lambda=0$. When implementing this idea, one needs to take
care of a few technical aspects: (1) as $\varepsilon \rightarrow 0$, the
denominator becomes small, hence something needs to be done to avoid amplifying
the errors; (2) if $x$ comes from the same dataset as the one used to train the
model, the model can be biased towards large absolute values of the log odds
on those samples, which can lead to biased estimates of the FIM; (3) $\vlambda$
can be multi-dimensional and we would like estimates of the FIM at all points of $\mcM$.

Based on the above idea, we propose ClassiFIM as an \emph{unsupervised} ML method to solve FIM-estimation tasks. It consists of three steps described below: transform, train, and estimate.

\subsection{Step 1: Transform the dataset}
The first step is to transform the input dataset
$\mcDtrain = \{(\vlambda_i, x_i)\}_{i=1}^{\abs{\mcDtrain}}$,
which has no labels, into a dataset $\mcDtrainBC = \{(\vlambda_0, \vdlambda, x, y)\}$
with labels $y$ suitable for training a BC. In every row, the first two components
$(\vlambda_0, \vdlambda)$ represent a pair of distinct
points $\vlambda_{\pm} = \vlambda_0 \pm \vdlambda/2$
in the parameter space $\mcM$, $y = \pm$ is a sign picked randomly with probability $0.5$ each
(the label for binary classification),
and $x$ is a random sample from the dataset $\mcDtrain$ with $\vlambda = \vlambda_{y}$.
This distribution is a substitute for the (unknown) distribution $P_{\vlambda=\vlambda_{y}}$.

This can be achieved, e.g., by \cref{alg:generateBC}, or a vectorized algorithm described in \cref{ass:fine-tune-alg1}.
In practice, the training data $\mcDtrainBC$ can be generated on-the-fly during training.
\begin{algorithm}
  \caption{Generating $\mcDtrainBC$ from $\mcDtrain$}
  \label{alg:generateBC}
  \textbf{Input}: dataset $\mcD$, even positive integer $N_{\textnormal{BC}}$\\
  \textbf{Output}: $\mcDBC$
  \begin{algorithmic}[1]
    \STATE $\mcDBC \leftarrow \varnothing$
    \WHILE{$\abs{\mcDBC} < N_{\textnormal{BC}}$}
    \STATE{Pick $(\vlambda_{+}, x_{+}), (\vlambda_{-}, x_{-}) \sim \mcD$}
    \COMMENT{two i.i.d. samples}
    \IF{$\vlambda_{+} = \vlambda_{-}$}
    \STATE{\textbf{continue}}
    \COMMENT{i.e., go to line 2}
    \ENDIF
    \STATE{$\vlambda_0 \leftarrow (\vlambda_{+} + \vlambda_{-})/2$}
    \STATE{$\vdlambda = \vlambda_{+} - \vlambda_{-}$}
    \STATE{$\mcDBC \leftarrow \mcDBC %
        \cup \{(\vlambda_0, \vdlambda, x_{-}, {-}), (\vlambda_0, \vdlambda, x_{+}, {+})\}$} %
    \COMMENT{the last component is the label}
    \ENDWHILE
    \STATE \textbf{return} $\mcDBC$
  \end{algorithmic}
\end{algorithm}

\subsection{Step 2: Pick and train a model for
  \texorpdfstring{$\mcDtrainBC$}{DtrainBC}}
\label{ss:bitchifc.step2}
We pick and train a BC model $\vecl = M(\vlambda_0, \vdlambda, x; \vectheta)$
satisfying the following conditions:
\begin{enumerate}
  \item[(1)] the output probability is given by
    $\hat{P}(Y = {+}1 | \vlambda_0, \vdlambda, x) = (1 + e^{-\vdlambda \cdot \vecl})^{-1}$;
  \item[(2)] the learned function
    $\vecl = M(\vlambda_0, \vdlambda, x; \vectheta)$
    extrapolates well as $\vdlambda \rightarrow 0$;
  \item[(3)] the estimated values of log odds $\vdlambda \cdot \vecl$ are
    accurate estimates of $\ln\left(P_{\vlambda_{+}}(x) / P_{\vlambda_{-}}(x)\right)$.
\end{enumerate}

\subsection{Step 3: Estimate the FIM}
\label{ss:bitchifc.step3}
After training this BC, one can estimate the FIM, $\hat{g}$, using the following formula:
\begin{equation}
  \label{eq:fim-from-nn-output}
  \hat{g}_{\mu\nu}(\vlambda) =
  \mean_{(\vlambda, x) \in \mcDtrain}(l_{x\mu}l_{x\nu}),
\end{equation}
where $\vecl_{x} = M(\vlambda, 0, x)$. Here, $(\vlambda, x) \in \mcDtrain$ means that the mean is taken with respect to all pairs $(\vlambda', x) \in \mcDtrain$ satisfying $\vlambda' = \vlambda$.

\subsection{Discussion}
Several comments about conditions (1)--(3) from step 2
and formula \cref{eq:fim-from-nn-output} are in order.
Condition (1) requires that log odds are computed as $\vdlambda \cdot \vecl$, which allows
us to avoid division by a small number in \cref{eq:fim-from-nn-output}.
Condition (2) is important because $\mcDtrainBC$ only contains rows with $\vdlambda \neq 0$,
but \cref{eq:fim-from-nn-output} evaluates $M$ at $\vdlambda = 0$.
While condition (3) may seem to be automatically satisfied by training
with cross-entropy loss as an objective, an overparameterized unregularized model
may memorize the training data and, e.g., find out that given a fixed finite $\mcDtrain$,
it is optimal to output large log odds even when $P_{\vlambda_{+}}(x) = P_{\vlambda_{-}}(x)$
(e.g., when $(\vlambda_{+}, x) \in \mcDtrain$ but $(\vlambda_{-}, x) \notin \mcDtrain$).

As demonstrated by our open-source implementation \citeImplementation{} and numerical
experiments in \cref{sec:numerical-experiments},
conditions (2) and (3) can be achieved in practice by training a BC with a small $L_2$ regularization,
i.e., training in the classical regime as opposed to the modern regime.
In particular, we do not use normalization layers since they change the role of weight decay
in the previous layers.

The challenge with using ClassiFIM with BC trained in the modern regime is the
tendency of the overparameterized models to achieve near-zero cross-entropy loss
on the training set (which is possible if all samples $x$ are distinct; see, e.g.,
\cite{zhang2021understanding,ishida2020we}), which leads to inflated absolute values of the log odds
and biased estimates $\hat{g}$ (i.e., typically, $\hat{g} > g$).

% TODO:8: Investigate ClassiFIM or test cross-entropy loss in the modern regime.
% TODO:7: Or, at least, provide an example plot of $\hat{g}$ trained in the modern regime
% in the appendix.

Our implementation is described in
\cref{sec:nn-design} and \cref{ss:classiFIM-BC-implementation}.

\subsection{Theoretical justification}
Here, we formally justify why ClassiFIM works, i.e., we formalize and prove the idea
that if the BC model is perfect, then \cref{eq:fim-from-nn-output} provides an unbiased
estimate of the FIM, which converges to the true FIM as the dataset size goes to infinity.

\begin{theorem}
  \label{th:bitchifc}
  Consider a statistical manifold-like pair $(\mcM, P)$ s.t. $\mcM \subset \mathbb{R}^m$
  and the space $\Omega$ of all possible samples $x$ is finite.
  Let $\vlambda_0$ be a point in the interior of $\mcM$, and $M^*$ be a model, satisfying
  the following conditions:
  \begin{enumerate}
    \item $M^*$ is optimal at $\vlambda_0$, i.e., for all $\vdlambda$ s.t.
      $\vlambda_{\pm} = \vlambda_0 \pm \vdlambda/2 \in \mcM$, and all $x \in \Omega$,
      $M^*(\vlambda_0, \vdlambda, x)$ satisfies
          \begin{equation}
            \label{aeq:th.bitchifc.1}
            \left(1 + e^{-\vdlambda \cdot M^*(\vlambda_0, \vdlambda, x)}\right)^{-1}
            \left(P_{\vlambda_+}(x) + P_{\vlambda_-}(x)\right) =
            P_{\vlambda_+}(x).
          \end{equation}
    \item The function
          $\vdlambda \mapsto M^*(\vlambda_0, \vdlambda, x)$ is continuous at $\vdlambda = \veczero$.
    \item $\forall x \in \Omega$ the function $\vlambda \mapsto P_{\vlambda}(x)$
          is differentiable at $\vlambda_0$.
    \item $\forall x \in \Omega$ with $P_{\vlambda_0}(x) = 0$ we have
          $P_{\vlambda_0 + \vdlambda}(x) = o\left(\norm{\vdlambda}^2\right)$.
  \end{enumerate}
  Then
  \begin{equation}
    \label{aeq:th.bitchifc.ps1}
    g_{\mu\nu}(\vlambda_0)
    = \mathbb{E}_{x\sim P_{\vlambda_0}}(l_{x\mu}^*l_{x\nu}^*),
    \quad\textrm{where}\quad
    \vecl_x^* = M^*(\vlambda_0, 0, x).
  \end{equation}
\end{theorem}

Here we assume that the definition of the FIM is extended
to cases where probabilities of individual events
might be equal to zero as explained in \cref{as:fim}.
The proof is given in \cref{as:bitstring-chifc}.

\section{Comparison with prior work}
\label{sec:comparison}

\begin{figure*}
  \includegraphics[width=\dimexpr0.99\textwidth\relax]{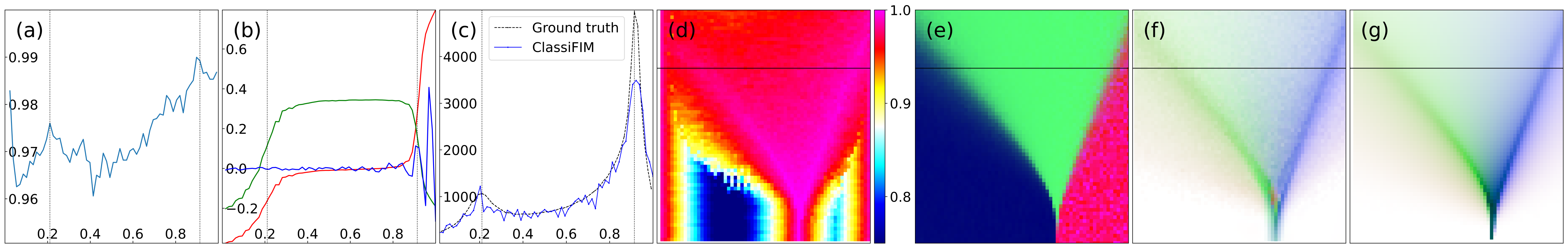}
  \caption{
    Illustration of the differences between the outputs of three unsupervised ML methods:
    W \cite[]{vanNieuwenburg2016LearningPT}, SPCA \cite[]{Huang:22},
    and ClassiFIM,
    when applied to the same dataset for IsNNN400 SM.
    Panels (a)--(c) show 1D phase diagrams generated by W, SPCA, and ClassiFIM, respectively,
    for $\lambda_1 = 48/64$ [the slice indicated by the black line in panels (d)--(g)].
    The vertical dotted lines indicate the maxima of the ground truth FIM shown in
    panels (c) and (g).
    Panels (d)--(f) show 2D phase diagrams generated by W, SPCA, and ClassiFIM,
    respectively.
    Panel (g) shows the ground truth FIM.
    Note that the scales on the diagrams (a)--(c), as well as color schemes on
    the diagrams (d)--(g),
    are different, reflecting the different meanings of the outputs of the methods: the
    W method produces the accuracy of ``mislabelled'' samples, which is expected to
    achieve a peak with a value close to $1.0$ at the locations of phase transitions;
    SPCA produces the components of a kernel principal components analysis (PCA) which are expected to change rapidly at the
    locations of phase transitions; ClassiFIM produces an estimate of the Fisher Information Metric,
    reflecting the rate of change of the underlying probability distribution.
    For more details see \cref{as:prior-work} (\cref{fig:isnnn400-w-details,fig:isnnn400-cf-details}).}
  \label{fig:isnnn400-1x7}
\end{figure*}

Phase transitions are often characterized by abrupt changes in the probability distribution
of the states of the physical system as the parameters $\vlambda$ change slightly.
Since the FIM represents the square rate of change of the underlying probability
distribution, peaks of the FIM estimated using the ClassiFIM method can be used
as a proxy for the locations of phase transitions.
Here, we describe two alternative methods for unsupervised learning of phase transitions,
which can be applied to the same datasets as ClassiFIM, and, thus, can be compared
to it.\footnote{%
  The authors have recently become aware of three additional methods
  that solve the FIM-estimation task and, therefore, can be directly compared to ClassiFIM:
  $I_2$ and $I_3$ from \cite{arnold2023machine}, and FINE introduced in \cite{duy2022fisher}.
  We plan to include the comparison with these methods in a future version of this work.}
The outputs of these two approaches and ClassiFIM,
when applied to one of the ten IsNNN400 datasets are illustrated in \cref{fig:isnnn400-1x7}.
IsNNN400 is based on a classical antiferromagnetic Ising model with nearest
and next-nearest neighbor (diagonal) interactions,
involving $400$ spins on a $20\times 20$ grid.

The first approach \cite[]{vanNieuwenburg2016LearningPT}, referred to as W below, computes a
``W''-like shape with the middle peak at the location of the phase transition.
There are three key challenges in comparing this method to ClassiFIM.
First, \cite{vanNieuwenburg2016LearningPT} suggests using the input dataset containing
so-called ``entanglement spectra'' for each $\vlambda$.
As noted in \citeCompanionPaper[App. K]{},
this data is hard to generate for larger system sizes,
either numerically or experimentally;
moreover, given this data, one can already predict a phase boundary without the use
of ML by directly computing the FIM,
which defeats the purpose of applying ML.
Second, the method is directly applicable only to one-dimensional phase diagrams
(i.e., $N=1$, where $N$ is the number of components of $\vlambda$).
Third, the method lacks a well-defined ground truth or a quantitative success metric
(there are no quantitative metrics
proposed in \cite{vanNieuwenburg2016LearningPT} to judge the quality of W plot; instead,
one can inspect it visually or compare the predicted location of phase transition with
the ground truth).

The second method \cite[]{Huang:22}, referred to as SPCA below,
is to apply kernel principal component analysis (PCA) to
\emph{classical shadows} which can be efficiently obtained on a quantum computer
capable of preparing the ground state of the system. \citeCompanionPaper[App. J]{}
has shown how to apply this method to the same dataset $\mcDtrain$ as ClassiFIM
instead of classical shadows.
The output of SPCA can be visualized by using the first 3 PCA components as color channels,
resulting in different colored regions that can be interpreted as phases.

Since W and SPCA do not solve the FIM-estimation task, we cannot use the metrics
$\distMSE$, $\distMSEPS$, and $\distRE$ to compare them to ClassiFIM.
Instead, to overcome the above-listed challenges,
we use the PeakRMSE metric introduced in \citeCompanionPaper[App. L]{},
which measures the accuracy of the predicted phase boundaries in the 1D slices
of the phase diagrams.
There is only one slice for 1D SMs but 128 slices
for 2D SMs described by a $64\times 64$ grid: $64$ vertical and $64$ horizontal.
Following the methodology of PeakRMSE, we need to modify all methods to
take the maximal number of guesses $n_s$ for each slice $s$ and output
$n_s$ guesses for a position of a phase-transition-like feature on that slice.
PeakRMSE is then given by
\begin{equation}
  \label{eq:peakrmse}
  \text{PeakRMSE} = \sqrt{\mean_{s,j} \min_k (x_{s,k} - y_{s,j})^2},
\end{equation}
where the list $\{x_{s,k}\}_{k=0}^{n_s+1}$ contains $n_s$ guesses and two boundary
points of the slice and $\{y_{s,j}\}_{j=1}^{n'_s}$ are the locations of the peaks
of the ground truth FIM on slice $s$, $n'_s \leq n_s$.
The guesses should be made with the goal of minimizing PeakRMSE.
We call such modified methods mod-W, mod-SPCA, and mod-ClassiFIM.

To ensure a fair comparison,
we attempt to adhere to the following principles for all methods:
(i) use the same input ($\mcDtrain$ and descriptions)
for making the predictions;
(ii) avoid any comparisons with ground truth until
the neural network (NN) architectures, hyperparameters,
and training procedures are finalized;
(iii) impose the same computational time limit;
(iv) include post-processing to improve the PeakRMSE.
We document in \cref{as:prior-work} cases where these rules were broken.
There are natural modifications for all three methods for the PeakRMSE metric.
In mod-W, we perform the training and extraction of the accuracy plot separately
for each slice and optimize the training time by training all $63$ networks
corresponding to a single slice in parallel. The accuracy plots are post-processed
using slope-capping. In both mod-W and mod-ClassiFIM,
the highest-prominence peaks (\verb!scipy.signal.find_peaks!)
are submitted as the guesses. Post-processing for mod-ClassiFIM and mod-SPCA
involves Gaussian smoothing. To extract peaks in mod-SPCA, we use k-means clustering on
the first two PCA components augmented to encourage spatial clustering.
For more details, see \cref{as:prior-work}.
We report the comparison results in the PeakRMSE column of \cref{tab:qpt-results}
in \cref{sec:numerical-phase-transitions-experiments} below.

\section{Numerical experiments}
\label{sec:numerical-experiments}
\subsection{Datasets and compute limits}
\label{ss:datasets}
The datasets used in our numerical experiments are summarized in \cref{tab:datasets}.
The first six rows are from \citeCompanionPaper[Table I]{} and
correspond to the six physical SMs (60 datasets) introduced there.
The last row corresponds to the ML dataset MNIST-CNN
introduced in \cref{ss:mnist-cnn-dataset}.
Each of the 61 datasets $\mcD$ has 140 samples for each $\vlambda \in \mcM'$
and is split into $90\%$ $\mcDtrain$ and $10\%$ $\mcDtest$.
The six physical SMs have a ground truth FIM, while MNIST-CNN does not.
When multiple datasets are available for the same SM,
they are effectively different only because of the different random seeds used
to generate the samples.
The values in the ``Time'' column are the suggested time limits in minutes
when executing on a system with a single NVIDIA GeForce GTX 1060 6GB GPU
(to be scaled down appropriately when training on more modern hardware).
We add a proposed value for MNIST-CNN and use these
to ensure a fair comparison between the methods (ClassiFIM, W, SPCA).

\begin{table*}[ht]
  \begin{center}
    \caption{\label{tab:datasets} Summary of statistical manifolds and datasets used in this work.}
    \begin{tabular}{cccccccccc}
      \toprule
      Name &$\#\mcD$ &$\#$features &$\mcM$ & $\abs{\mcM'}$ &Time &system description \\
      \midrule
      Ising400& 10 &$400$& $(0, 1]$   & 1000 & 10 & 2D classical, Gibbs state  \\
      IsNNN400& 10 &$400$& $(0, 1]^2$ & 4096 & 20 & 2D classical, Gibbs state  \\
      FIL24   & 10 & $24$& $[0, 1)^2$ & 4096 & 10 & 2D Spin-1/2, ground state  \\
      Hubbard12&10 & $24$& $[0, 1)^2$ & 4096 & 10 & 2D Fermionic, ground state \\
      Kitaev20& 10 & $20$& $(-4, 4)$  &20000 & 10 & 1D Fermionic, ground state \\
      XXZ300  & 10 &$300$& $[0, 1]^2$ & 4096 &125 & 1D Spin-1/2, ground state  \\
      MNIST-CNN& 1 &$795$& $[0, 1)^2$ & 4096 & 60 & outputs of neural networks \\
      \bottomrule
    \end{tabular}
  \end{center}
\end{table*}

\subsection{MNIST-CNN dataset}
\label{ss:mnist-cnn-dataset}
We construct the MNIST-CNN dataset as a toy example to show how ClassiFIM can be used to study the space of different regimes that occur in the training of a neural network. Many works (see \cite{baldassi2022learning} and references therein) interpreted regimes such as success and failure of learning, as well as highly overparametrized behavior, as distinct phases of matter of the dynamical system that is the neural network, with the state given by its weights and/or the table of predictions. We note that while this pursuit did not directly influence any performance improvement, it is of academic interest nonetheless. Here we suggest repurposing MNIST for such investigations, while deferring the majority of the required analysis to future work.
The MNIST-CNN SM is based on a CNN model \cite{tuomaso2021mnist} for the MNIST dataset \cite[]{lecun1998mnist}. With specific choices of the maximum learning rate $\mathrm{max\_lr}=10^{-2}$ used in OneCycleLR \cite[]{smith2019super} and the parameter $\beta_1=0.7$ used in the Adam optimizer, the network achieves 99\% accuracy on MNIST in just one epoch ($\approx 1$ second), but performance can degrade dramatically away from the optimal choice.

The parameter space $\mcM$ of the MNIST-CNN dataset is a set of pairs $(\lambda_0, \lambda_1) \in [0, 1)^2$,
where $\lambda_0 = (\log_{10}(\mathrm{max\_lr}) + 3) / 2$,
$\lambda_1 = -\log_{10}(1-\beta_1)/4$. The grid $\mcM'$ consists of points $\vlambda$, where $\lambda_j = k_j/64$ for $k_j=0,\dots,63$, $j=0,1$. For each $\vlambda$, a CNN is trained with the corresponding parameters. In each row $(\vlambda, x)$ of the dataset $\mcDtrain$, the sample $x = (i, \hat{p}, y)$ contains
(i) a randomly chosen image $i$ from MNIST,
(ii) the output class probabilities $\hat{p}$ for image $i$ returned by the CNN trained with the parameters $\vlambda$,
(iii) the ground truth MNIST label $y$ assigned to $i$.

Each of the 4096 trained CNNs has two sources of randomness: the weights are pseudorandomly initialized, and the training dataset is a pseudorandom permutation of the MNIST training dataset.
Given this randomness, although we trained only one model for each $\vlambda$, we conceptualize the dataset as representing a SM where each sample $x$ is obtained from a randomly initialized CNN evaluated on a randomly chosen image.
The dataset's samples have the correct probabilities $P_{\vlambda}(x)$,
but the samples $x$ are not independent (making them independent would require training $140 \times 4096$ instead of $4096$ CNNs).
Unlike the physics datasets, we do not provide the ground truth FIM because, in the MNIST-CNN case, it is unclear how to generate it.

\subsection{Neural network design and implementation}
\label{sec:nn-design}
\begin{figure}
  \includegraphics[width=\dimexpr0.998\textwidth\relax]{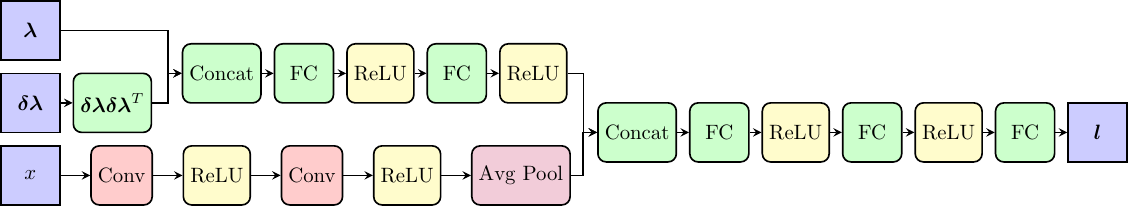}
  \caption{
    \label{fig:nn-architecture}
    The neural network architecture we used for the Hubbard12 and FIL24 datasets.
    Here ``Conv'' layers are graph convolutional layers for the $12$-site lattice
    with two different types of edges.}
\end{figure}

To implement ClassiFIM on the seven SMs,
following the prescription given in \cref{ss:bitchifc.step2},
we implement six neural network (NN) architectures
reflecting the format of the input data:
one for each SM with the exception of FIL24, where we use the
same NN and training pipeline as for Hubbard12.
For example, the Hubbard12 NN
architecture is depicted in \cref{fig:nn-architecture}.
% Fact check:
% * Hubbard12 and FIL24 use Pipeline in twelve_sites_pipeline.py, which uses
%   TwelveSitesNN2 model from layers.py matching \cref{fig:nn-architecture}.
% * IsNNN400 uses Pipeline in ising2d, which uses ModelNaiveCNN. That model
%   has only 1 CNN layer, but one more FC layer.
Neural networks for other datasets mainly
differ in the type of convolutional layers used, number of layers, the number of channels.
The type of convolutional layer used is chosen based on the structure of the data:
e.g., 1D CNN for 1D data, 2D CNN for 2D data, with padding matching the boundary conditions
of the system. Other hyperparameters of NN architecture together with the number of epochs
used during the training are adjusted to match the time limit given in \cref{tab:datasets}.
Each NN is trained to minimize the cross-entropy error
for the auxiliary binary classification task,
and includes a small $L_2$ regularization on the network weights.
The code with the exact neural network architecture, training procedure, and
learned hyperparameters used for each SM is provided in \citeImplementation{}.
These NN architectures and hyperparameter choices are based on early experiments
on lower-quality datasets not described in this work. We use the Adam optimizer
\cite{kingma2014adam}
and a OneCycleLR learning rate scheduler \cite{smith2019super}.

\subsection{Comparison of metrics}
\label{ss:comparison-of-metrics}
Here we evaluate how well the metrics $\distMSE$, $\distMSEPS$, $\distRE$, and PeakRMSE
can distinguish between the performance of two solutions to the FIM-estimation task.
In order to do this, we trained ClassiFIM on the XXZ300 datasets using all available time,
which we name CF580 since we could train it for $580$ epochs,
and compare it to a model trained for $200$ epochs, which we name CF200.
Such a comparison is repeated for $10$ datasets, allowing us to perform a paired t-test
to determine if the difference in the performance of the two models is
statistically significant. As one can see in \cref{tab:metrics-comparison},
each of the 3 metrics $\distMSE$, $\distMSEPS$, and $\distRE$
is able to determine that CF580 is better than CF200, but PeakRMSE cannot.
Intuitively, this is because PeakRMSE is based only on a few
properties of the prediction (locations of 126 predicted peaks),
while the other metrics take into account all $3 \times 64^2$ predicted values
$\hat{g}_{\mu\nu}(\vlambda)$.
\begin{table}
  \caption{Comparison of metrics for two ClassiFIM models trained on the XXZ300 datasets for different numbers of epochs. The values in the table are the mean values of the metrics for the two models, and the $p$-values are obtained from a paired t-test.
    In columns CF200 and CF580 and in \cref{tab:qpt-results} values in brackets indicate the (rounded) standard deviation (across the $10$ datasets) in units corresponding to the last place of the value (omitted when $(0)$):
  e.g., $4.5(4)$ indicates a standard deviation of $\approx 0.4$, and $26.06$ indicates a standard deviation less than $0.005$.}
  \centering
  \begin{tabular}{llll}
    \toprule
    Metric      & CF200    & CF580    & $p$-value \\
    \midrule
    PeakRMSE    & $0.0080(7)$ & $0.0078(5)$ & $0.536$ \\
    $\distMSE$  & $0.78(17)$  & $0.39(7)$  & $2.2 \cdot 10^{-4}$ \\
    $\distMSEPS$& $0.57(9)$  & $0.36(4)$  & $4.6 \cdot 10^{-5}$ \\
    $\distRE$ ($\%$)& $1.42(11)$ & $1.27(7)$  & $3.2 \cdot 10^{-3}$ \\
    \bottomrule
  \end{tabular}
  \label{tab:metrics-comparison}
\end{table}
By definition,
the three metrics are invariant under linear transformations of the parameter space,
while PeakRMSE is not.

\subsection{ClassiFIM performance on physics datasets}
\label{sec:numerical-phase-transitions-experiments}

\begin{table}
  \caption{Performance of ClassiFIM (CF) on physics datasets. CE = cross-entropy. $\distRE$ values are in $\%$.
  The best method(s) according to the PeakRMSE column are highlighted in \textbf{bold} (multiple values are highlighted when the difference is not statistically significant).
  Methods in the PeakRMSE part refer to their modified versions
  (mod-ClassiFIM, mod-SPCA, and mod-W).}
  \centering
  \begin{tabular}{llllllll}
    \toprule
    SM & Method & Test CE & $\distMSE$  & $\distMSEPS$ & $\distRE$ & Method & PeakRMSE \\
    \cmidrule(lr){1-1} \cmidrule(lr){2-6} \cmidrule(lr){7-8}
    IsNNN & CF    &$0.0755(2)$& $11(1)$   & $4.6(9)$ & $4.5(4)$ & CF   &$\mathbf{0.017}(4)$\\
    400 & const &$0.6931$   & ---       &$55.6(2)$ & $26.06$  & SPCA &$\mathbf{0.016}(1)$\\
          &       &           &           &          &          & W    & $0.035(7)$ \\
    \cmidrule(lr){1-1} \cmidrule(lr){2-6} \cmidrule(lr){7-8}
    Hubbard & CF    &$0.250(3)$ & $4.0(6)$  & $1.5(1)$ & $7.7(6)$ & CF   &$\mathbf{0.035}(7)$\\
    12 & const &$0.6931$   & ---       & $6.2032$ & $23.64$  & SPCA &$\mathbf{0.039}(1)$\\
     & best  &$0.2380(9)$& $0.0000$  & $0.0000$ & $0.000$  & W    & $0.076(2)$ \\
    \cmidrule(lr){1-1} \cmidrule(lr){2-6} \cmidrule(lr){7-8}
    FIL24 & CF    &$0.2422(8)$& $0.47(3)$ & $0.28(1)$& $4.4(1)$ & CF   & $0.028(8)$ \\
          & const &$0.6931$   & ---       & $4.2078$ & $25.60$  & SPCA &$\mathbf{0.017}(1)$\\
          & best  &$0.2410(8)$& $0.0000$  & $0.0000$ & $0.000$  & W    & $0.026(4)$\\
    \cmidrule(lr){1-1} \cmidrule(lr){2-6} \cmidrule(lr){7-8}
    XXZ300   & CF    &$0.0945(1)$& $0.39(7)$ & $0.36(4)$& $1.27(7)$& CF  &$\mathbf{0.0078}(5)$\\
      & const &$0.6931$   & ---       &  $41.307$& $24.55$  & SPCA & $0.012(2)$\\
          &       &           &           &          &          & W    & $0.011(3)$\\
    \cmidrule(lr){1-1} \cmidrule(lr){2-6} \cmidrule(lr){7-8}
    Kitaev20 & CF    &$0.2086(7)$&$5.4(1)$   & $0.71(2)$& $3.2(1)$ \\%& CF   & \\
          & const &$0.6931$   & ---       &$1.6873(6)$&$7.21$   \\%& SPCA & \\
    %      &       &           &           &          &          & W    & \\
    \cmidrule(lr){1-1} \cmidrule(lr){2-6} % \cmidrule(lr){7-8}
    Ising400 & CF    &$0.1745(4)$& $0.69(10)$&$0.55(6)$& $1.52(7)$ \\%& CF   & \\
       & const &$0.6931$   & ---       &$3.58(6)$ & $10.26$  \\%& SPCA & \\
    %      &       &           &           &          &          & W    & \\
    \bottomrule
  \end{tabular}
  \label{tab:qpt-results}
\end{table}
The results of our numerical experiments on the physics datasets
are summarized in \cref{tab:qpt-results}. This table comprises six sections,
each corresponding to one SM, for which we have $10$ datasets.
The ``Test CE'' column compares the hold-out ($\mcDtest$) cross-entropy for the auxiliary binary
classification task
of the ClassiFIM model we have trained with a ``const'' model (returning 0 log odds)
and, where available, the best possible model returning the ground truth log odds
for that SM.
The $\distMSE$, $\distMSEPS$, and $\distRE$ columns
show the metrics for the FIM-estimation task.
Here the row labeled
``const'' corresponds to predicting the constant FIM corresponding to the Euclidean metric on the parameter space, with an unknown scaling factor ($\distMSE$ metric is not scale-invariant and, thus, is undefined for the ``const'' method)
and ``best'' corresponds to predicting the ground truth FIM.
For more details on ``const'' and ``best'' rows see \cref{as:comparison}.
We find that the ClassiFIM model is more precise on the easier FIL24 SM
even though we reused the same architecture,
hyperparameters, and training procedure as were used on the Hubbard12 SM.
Apart from the metrics in \cref{tab:qpt-results}, the quality of the agreement
between the ClassiFIM prediction and the ground truth on one of the ten IsNNN400 datasets
can be seen by comparing panels (f) and (g) of \cref{fig:isnnn400-1x7}
or the two curves in panel (c).

The last two columns of \cref{tab:qpt-results} compare the performance of
ClassiFIM with other state-of-the-art methods
(all modified as explained in \cref{sec:comparison}).
We note that all methods, including ClassiFIM, used, on average, $126$ examples
per $\vlambda$, which is a fixed amount (independent of the lattice size of
the underlying physical system), and is significantly less than the number of
probabilities needed to compute the FIM directly from the definition
(e.g., $2^{300}$ for XXZ300).
In summary, from the perspective of this metric,
mod-ClassiFIM outperforms mod-W and is competitive with mod-SPCA.

\subsection{Use of ClassiFIM on MNIST-CNN dataset}
\label{ss:mnist-cnn-results}

\begin{figure}
  \centering
  \includegraphics[width=0.24\linewidth]{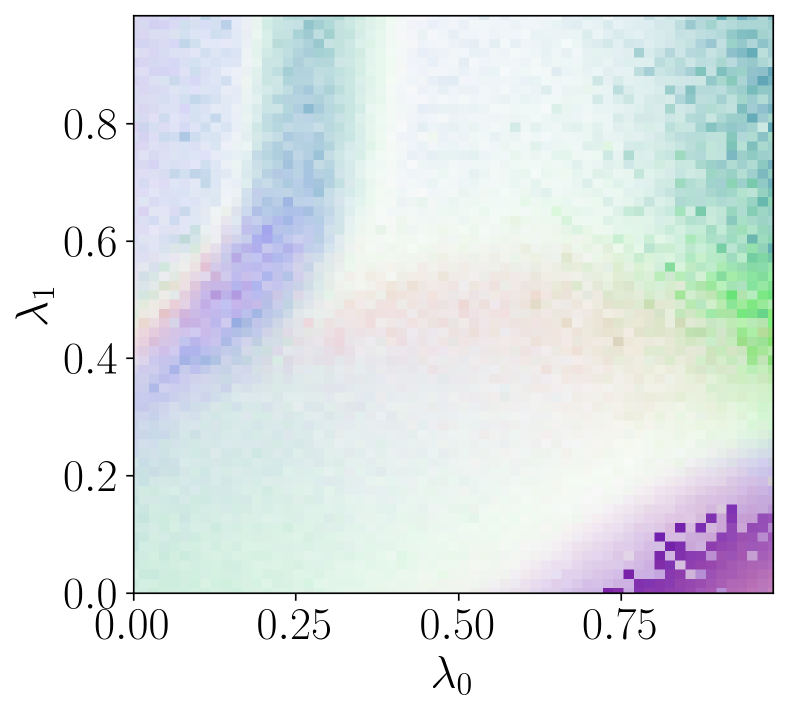}
  \includegraphics[width=0.74\linewidth]{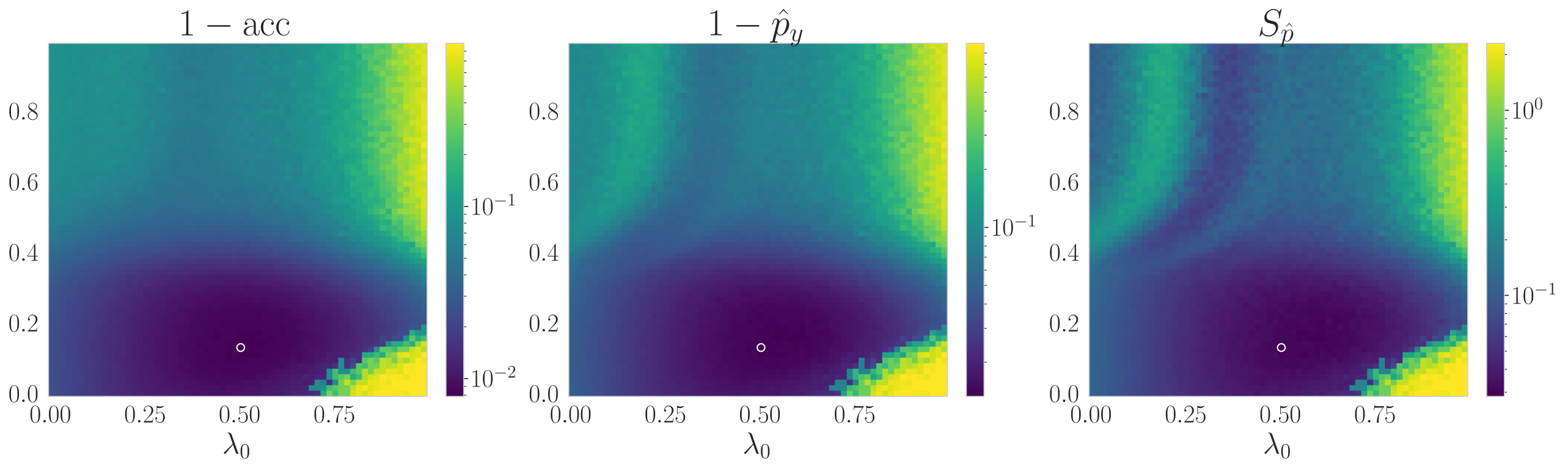}
  \caption{
    \label{fig:MNIST}
    ClassiFIM output (left) compared with properties of the outputs of the models
    trained on MNIST.
    From left to right, these are $1 - \mathrm{accuracy}$, $1 - \mathbb{E}\hat{p}_y$,
    where $\hat{p}_y$ is the predicted probability of the correct class,
    and the mean entropy of the predicted class probabilities.
    For these three plots, the averages are computed over all images in the training set,
    and over $14$ trained models for each $\vlambda$.
    The axes correspond to $\mathrm{max\_lr} = 10^{-3+2\lambda_0}$
    and $\beta_1 = 1 - 10^{-4 \lambda_1}$. The white circle indicates the $\approx 99\%$
    accuracy model from \cite{tuomaso2021mnist}.
  }
\end{figure}

We present the results of using ClassiFIM on MNIST-CNN in \cref{fig:MNIST}.
The FIM phase diagram detects rapid changes in the behavior of the model in the top and bottom right regions, which coincide with the loss of accuracy. A region with many sharp peaks in the FIM may also be observed in so-called glassy systems (see, e.g., Figs. 3 and 4 of \cite{takahashi2019phase}), for which it may be possible to obtain the ground truth and verify the performance of ClassiFIM. \cref{fig:MNIST} also suggests a phase transition in the upper-left corner:
a stripe darker than the surrounding area indicates a peak in the predicted FIM.
This boundary is not captured by accuracy alone: there is no sign of a rapid
change in the accuracy plot (second panel from the left) at that location. However, one can find other
quantities that are capable of capturing this transition: it can be seen
in the mean probability of the correct class, $\mathbb{E}\hat{p}_y$ (third panel),
and, more clearly, in the mean entropy of the predicted class probabilities (last panel). Since we did not study ClassiFIM in the glassy phase, we do not have high confidence in how to interpret the predicted high values shown in the ClassiFIM output in \cref{fig:MNIST}, and cannot exclude a potential mis-prediction. We leave studying the FIM and ClassiFIM for phase diagrams having a transition to a glassy phase as a challenging future direction. 
We note that while the right three panels of \cref{fig:MNIST} use averages over the train split of MNIST,
the plots for the test split are qualitatively similar.

\subsection{Scalability}
\label{ss:scalability}
For datasets of similar subjective complexity (Ising400 and IsNNN400 are ``easier'' than QPT datasets in the next four rows), the time allocated (as prescribed by column ``Time'' in \cref{tab:datasets}) is roughly proportional to the size of input data, i.e., $\abs{\mcDtrain} n$ where $n$ is the number of features in each sample $x_i$.
With that scaling, ClassiFIM achieves better performance on larger datasets of similar nature, both in terms of absolute values of $\distRE$ (which is invariant not only to the scale of the predicted FIM, but also to the scale of ground truth), and in terms of a comparison with other methods using PeakRMSE. This suggests that linear scaling of computational resources with input size suffices for the types of datasets considered here.

\section{Limitations and future work}
\label{s:limits-and-future}

While we have shown that ClassiFIM can solve the FIM-estimation task, we expect that there are FIM-estimation tasks where ClassiFIM or any other method would fail. Moreover, all of the neural networks we implemented in this work effectively used local features, which is incompatible with solving some otherwise solvable FIM-estimation tasks. Nevertheless, ClassiFIM does not require any particular choice of NN architecture, and future work could explore other choices and applications of ClassiFIM to other problems. See \cref{as:limitations} for an extended discussion of these topics.

%% START: embedded bu1.bib

%% END: embedded bu1.bib
\end{bibunit}

\newpage

\renewcommand*{\HyperDestNameFilter}[1]{#1-bu2}
\begin{bibunit}[abbrvnat] % The optional arg is the style

\appendix

\begin{center}
  {}\textbf{\large{Technical Appendix}}
\end{center}

Here, we provide additional details in support of the main text.
In \cref{as:fim}, we provide an additional discussion of the Fisher information metric.
In \cref{as:ds-description} we give an example of a textual description accompanying $\mcD$.
In \cref{as:benchmarking} we provide complete definitions of $\distMSE$, $\distMSEPS$, and $\distRE$.
In \cref{ass:classifim} we provide additional details about ClassiFIM: we prove \cref{th:bitchifc} (theoretical justification of ClassiFIM), describe the design choices we made in our implementation of ClassiFIM, and discuss an alternative algorithm for generation of $\mcDtrainBC$.
In \cref{as:prior-work} we describe our implementation of mod-W, mod-SPCA, and mod-ClassiFIM and discuss places were we deviated from rules we introduced in \cref{sec:comparison}.
In \cref{as:comparison} we define the ``const'' and ``best'' rows from \cref{tab:qpt-results} and discuss the naive method for FIM estimation.
Finally, in \cref{as:limitations} we discuss the limitations and future work previously outlined in \cref{s:limits-and-future}.

Other details are provided in the companion paper \citeCompanionPaper{}.
For example, the definitions of the physics-inspired statistical manifolds
used in our work are described in \citeCompanionPaper[Sec.~IV.B.]{};
the numerical details that went into generating the ground truth FIM
and samples for our datasets are described in \citeCompanionPaper[App.~H]{}.
The code can be found at \citeImplementation{} and the data at
\cite{public-datasets}.

\section{Background on Fisher Information Metric}
\label{as:fim}
In \cref{sec:fim}, \cref{eq:fim} we introduced the Fisher Information Metric (FIM).
In this section, for the convenience of the reader, we
reproduce the rigorous definition of the Fisher Information Metric (FIM)
and motivation behind it, as introduced in \citeCompanionPaper[App B]{}.
This definition is used in \cref{th:bitchifc}.

Given the formula
\begin{equation}
  s_{\mu}(x;\vlambda) = \partial{\lambda_{\mu}} \log P_{\vlambda}(x)
\end{equation}
for the score function, one might think that
four conditions are necessary for the FIM $g_{\mu\nu}(\vlambda)$ to be defined:
(i) the space of samples should form a disjoint union of one or more manifolds;
(ii) for each $\vlambda$ the probability distribution should be absolutely
continuous on each of those manifolds, so that the density function
$P_{\vlambda}(\bullet)$ is well-defined;
(iii) $P_{\vlambda}(x) \neq 0$ for all $x$;
(iv) $P_{\vlambda}(x)$ is differentiable with respect to $\vlambda$ for all $x$.
However, a more general treatment of the FIM is possible, which does not require
any of these conditions.
In fact, there are multiple inequivalent choices for
such a more general treatment (they become equivalent when the above conditions
are satisfied), and we select one of them based on our main
application.
These choices are based on the fact that the FIM
coincides (under certain regularity conditions,
up to a constant factor) with the first non-trivial term in the
Taylor expansion of multiple measures of distance or similarity between probability
distributions, including the Kullback-Leibler
divergence, Jensen-Shannon divergence, and Bhattacharyya distance
\cite[section 3.5 Fisher Information: The Unique Invariant Metric]{amari2016information}.
Specifically, we pick the last of these measures, the Bhattacharyya distance,
which is also known as the classical fidelity $F_c$ in the physics literature on quantum
phase transitions (but some sources define $F_c$ to be the square of r.h.s. of
\cref{eq:Fc.def} instead, see e.g., \cite[Def 9.2.4 on page 271]{wilde2011classical}).

Assume that we have a manifold $\mcM$ of parameters $\vlambda$ and a measurable space
$(\Omega, \mathcal{F})$, where $\Omega$ is the set of all possible samples, and
$\mathcal{F}$ is a sigma-algebra on that space. Furthermore, assume that for every
$\vlambda\in\mcM$ we have a probability measure $\sigma_{\vlambda}$.
The classical fidelity between two probability measures $\sigma$ and $\sigma'$ is
defined as
\begin{equation}
  \label{eq:Fc.def}
  F_c(\sigma, \sigma') = \int_{\Omega} \sqrt{d\sigma d\sigma'}.
\end{equation}
We note that $F_c$ is always well-defined, where the integral on the r.h.s. is
defined using the Radon-Nikodym derivative with respect to any measure with respect
to which both $\sigma$ and $\sigma'$ are absolutely continuous (e.g., $\sigma + \sigma'$). When $F_c(\sigma_{\vlambda_0}, \sigma_{\vlambda_0 + \vdlambda})$ has the form
\begin{equation}
  \label{eq:Fc.Taylor}
  F_c(\sigma_{\vlambda_0}, \sigma_{\vlambda_0 + \vdlambda}) =
  1 + \frac{1}{8} \sum_{\mu,\nu} g_{\mu\nu}(\vlambda_0)
  \delta\lambda_{\mu} \delta\lambda_{\nu}
  + o(\norm{\vdlambda}^2)
\end{equation}
for some $g_{\mu\nu}(\vlambda_0)$,
we say that the FIM is defined. In that case, we can always make $g_{\mu\nu}(\vlambda_0)$ symmetric with respect to the interchange of indices $\mu$ and $\nu$ by replacing it with $(g_{\mu\nu}(\vlambda_0) + g_{\nu\mu}(\vlambda_0))/2$. Then, such a symmetric $g_{\mu\nu}(\vlambda_0)$ is the FIM.

\section{Description of Hubbard12 and FIL24 datasets}
\label{as:ds-description}
As mentioned in \cref{ss:fim-estimation-task}, the datasets are accompanied
by an additional textual description. Here we provide such a description for the
Hubbard12 and FIL24 datasets (the description for these 20 datasets is identical):

\begin{quote}
  The twelve-site grid consists of $12$ sites indexed by $j \in \{0, 1, \ldots, 11\}$. There are two types of edges: NN and NNN. Each site $j$ is connected via NN edges with 4 sites: $(j \pm 1) \bmod 12$ and $(j \pm 3) \bmod 12$, and via NNN edges with 4 sites: $(j \pm 2) \bmod 12$ and $(j \pm 4) \bmod 12$. Each sample $x_i$ is an integer in $[0, 2^{24})$ representing a $24$-bit bitstring (with $2$ bits per lattice site). The bits of $x_i$ corresponding to site $j$ are $\floor{x_i / 2^{j}} \bmod 2$ and $\floor{x_i / 2^{12 + j}} \bmod 2$. These two bits may have different meanings.

  All NN edges are identical to each other. All NNN edges are identical to each other. All lattice sites are identical to each other. In other words, any graph automorphism of the twelve-site grid which preserves edge labels ``NN'' and ``NNN'' (e.g. $j \mapsto (j + k) \bmod 12$ or $j \mapsto 11 - j$) preserves the probability distribution over bitstrings $x_i$.
\end{quote}
Textual descriptions for all datasets are available in \cite{public-datasets}.
% TODO:6: verify it is actually there
While in this work we manually constructed NN architectures corresponding to such
descriptions, we hope that future work could automate this process, e.g., by
using an LLM to generate an NN architecture from a textual description of the dataset.

\section{Metrics and benchmarking}
\label{as:benchmarking}

\subsection{Performance metrics}
\label{ass:performance-metrics}

Let $g(\vlambda)$ denote the ground-truth FIM and $\hat{g}(\vlambda)$ denote an estimate. A seemingly reasonable performance metric is given by
\begin{equation}
  \label{app-eq:naive-error}
  \textrm{distNaive}(\hat{g}, g) = \int_{\mcM}\norm{g(\vlambda) - \hat{g}(\vlambda)}_{F}^2 d\vlambda,
\end{equation}
where $\norm{\bullet}_F$ is the Frobenius norm (recall that $g(\vlambda)$ is a rank-2 tensor). However, we are not aware of a natural geometric interpretation of $\textrm{distNaive}$ and
such a metric is not invariant with respect to parameterization change in $\mcM$. Another disadvantage of this metric is that it may assign a large error to a blurred estimator $\hat{g}_1$ when the ground truth FIM has sharp peaks, to the extent of making it worse than an estimator $\hat{g}_2$ missing half of the peaks (phase transitions) but otherwise matching the ground truth. This is inconsistent with the goal of qualitatively reproducing a phase diagram. To overcome some of these issues, we propose three metrics: $\distMSE$, $\distMSEPS$, and $\distRE$ which favor $\hat{g}_1$ over $\hat{g}_2$.

Given a Riemannian metric, a natural quantity to compute is the distance between two points along the shortest path in that metric. However, this approach is computationally expensive, and the geodesic connecting two points may transcend the region of
interest in the parameter space. As a practical alternative, we instead consider the ``straight line distance'' $\distSL$ between two points $\vlambda$ and $\vlambda'$:
\begin{equation}
  \label{app-eq:distSL}
  \distSL(g; \vlambda, \vlambda')
  = \int_0^1 dt \sqrt{
    g(\vlambda + t(\vlambda' - \vlambda); \vlambda' - \vlambda)},
\end{equation}
which we use to define our three metrics. Here $g(\vlambda; \vecv) = \sum_{ij} g(\vlambda)_{ij} v_i v_j$
is the square length of the vector $\vecv$ in the metric
$g$ at the point $\vlambda$. In our numerical experiments, both the ground truth $g$
and the prediction $\hat{g}$ are represented by their values on a grid $\mcM'$,
in which case we assume that these values represent a piecewise constant function.
Hence, \cref{app-eq:distSL} is an integral of a piecewise constant function and can be written as a sum. For a 1D parameter space, $\distSL$ for $N_{\mathrm{SL}}$ pairs $(\vlambda, \vlambda')$ can be computed in time $\Theta(N_{\mathrm{SL}} + \abs{\mcM'})$ using dynamic programming.
For grids in multi-dimensional parameter spaces (with a fixed number of dimensions), the cost is $O(N_{\mathrm{SL}} L)$ where $L$ is the largest grid dimension.

The first step in the computation of all 3 metrics is the computation of $\distSL(\bullet; \lambda, \lambda')$ for a large set of pairs $(\lambda, \lambda')$ of different points (e.g., all pairs of points in the grid $\mcM'$).
The first metric is the MSE for $\distSL$ given by,
\begin{equation}
  \label{app-eq:distMSE}
  \distMSE(\hat{g}, g)
  = \mean_{\vlambda \neq \vlambda' \in \mcM'}
  \bigl(
  [\distSL(\hat{g}; \vlambda, \vlambda')
    - \distSL(g; \vlambda, \vlambda')
  ]^2\bigr).
\end{equation}

A scale-invariant version $\distMSEPS$ is defined by optimizing the scaling constant:
\begin{equation}
  \label{app-eq:distMSEPS}
  \distMSEPS(\hat{g}, g) = \min_{c>0}(\distMSE(c\hat{g}, g))
\end{equation}

Finally, the third metric is the distance ranking error $\distRE$ representing the
probability that $\distSL(\hat{g}; \bullet, \bullet)$ ranks two pairs
of points differently from $\distSL(g; \bullet, \bullet)$. The exact definition of $\distRE$ is given below.
Out of the three metrics, $\distMSE$ penalizes for any mismatch between the prediction and the ground truth, $\distMSEPS$ allows for a (multiplicative) bias which may appear due to a lack of accurate calibration in otherwise accurate predictions, and $\distRE$ is the most lenient (it could be $0$ even if errors other than overall multiplicative errors occur which do not affect distance rankings).

\subsection{Definition of the distance ranking error}
\label{as:ranking-error}

When following the above-given recipe for defining $\distRE$, special care should be taken in the case of a tie. Here, we provide the complete definition.
The distance ranking error of $\hat{g}$ (compared to the ground truth
${g}$) is defined as
the average value returned from the following procedure:
\begin{itemize}
  \item Pick random $\vlambda_0, \vlambda_1, \vlambda_2, \vlambda_3 \in \mathcal{M}'$ s.t.
        \begin{itemize}
          \item $\vlambda_0 \neq \vlambda_1$ and $\vlambda_2 \neq \vlambda_3$, and
          \item the set (i.e., unordered pair) $\{\vlambda_0, \vlambda_1\}$
                is different from $\{\vlambda_2, \vlambda_3\}$.
        \end{itemize}
  \item Let $\hat a, a \in \{-1, 0, 1\}$
        be the results of a comparison of
        $\distSL(\hat{g}; \vlambda_0, \vlambda_1)$
        with $\distSL(\hat{g}; \vlambda_2, \vlambda_3)$ and of
        $\distSL({g}; \vlambda_0, \vlambda_1)$
        with $\distSL({g}; \vlambda_2, \vlambda_3)$, respectively:
        $-1,0,1$ correspond to $<,=,>$, respectively.
  \item If $a = 0$, return $0.5$. Otherwise, return $\abs{\hat a - a}/2$
        (e.g., $0.5$ if $\hat a = 0$).
\end{itemize}
Note that we introduced a special case for $a = 0$ to avoid the situation in which a method can decrease its error by rounding the values of $\hat{g}$
(to increase the probability of ties). This is equivalent to resolving the
ties at random (with a probability of $0.5$ for each outcome).

The metric $\distRE$ thus defined is symmetric with respect to $\hat{g}$ and
${g}$, and is always in $[0, 1]$. Using a merge-sort based inversion
counting algorithm, $\distRE$ can be computed in
$\Theta(N_{\mathrm{SL}} \log \abs{N_{\mathrm{SL}}})$ time,
where $N_{\mathrm{SL}}$ is the number of pairs for which $\distSL$ was computed
for (e.g., $N_{\mathrm{SL}} = \abs{\mcM'}(\mcM' - 1)/2$ if all values of $\distSL$ are used), which is typically less expensive than computing those $\distSL$ values in the first place.

\section{ClassiFIM method details}
\label{ass:classifim}

\subsection{Theoretical justification}
\label{as:bitstring-chifc}

\begin{proof}[Proof of \cref{th:bitchifc}]
  First, note that the r.h.s. of \cref{aeq:th.bitchifc.ps1} [$g_{\mu\nu}(\vlambda_0)
    = \mathbb{E}_{x\sim P_{\vlambda_0}}(l_{x\mu}^*l_{x\nu}^*)$, where $\vecl_x^* = M^*(\vlambda_0, 0, x)$] is
  well-defined and is symmetric with respect to interchanging
  $\mu$ and $\nu$. Let us define its value along some vector
  $\vecv$ as $\hat{g}(\vlambda_0;\vecv)$:
  \begin{equation}
    \label{eq:th.bitchifc.proof1}
    \hat{g}(\vlambda_0;\vecv) = \\
    \mathbb{E}_{x\sim P_{\vlambda_0}}\left(
    \sum_{\mu\nu} l_{x\mu}^*l_{x\nu}^* v_\mu v_\nu
    \right).
  \end{equation}
  Based on the definition we use given by \cref{eq:Fc.def,eq:Fc.Taylor},
  it is sufficient to show that, as $\epsilon \to 0$,
  \begin{equation}
    \label{eq:th.bitchifc.proof2}
    \frac{8}{\epsilon^2}
    \left(1-\sum_{x} \sqrt{P_{\vlambda_0}(x) P_{\vlambda_0 + \vecv \epsilon}(x)}\right)
    = \hat{g}(\vlambda_0;\vecv) + o(1).
  \end{equation}
  Indeed, if we show this for an arbitrary fixed $\vecv$, then we can write $\vdlambda$ from
  \cref{eq:Fc.Taylor} as $\epsilon \vecv$ with $\epsilon = \norm{\vdlambda}$
  and $\vecv = \vdlambda/\epsilon$. Then, it would remain to show that
  the remainder term $o(\epsilon^2)$ in \cref{eq:Fc.Taylor} is
  uniform in $\vecv$ with $\norm{\vecv} = 1$, which is necessarily the
  case because the unit sphere is compact and all terms are continuous in $\vecv$.

  First, let us try to replace the l.h.s. of \cref{eq:th.bitchifc.proof2},
  $f_{1/2}(\epsilon)$,
  with its symmetric version $f(\epsilon) = f_0(\epsilon)$, where
  \begin{equation}
    \label{eq:th.bitchifc.proof3}
    f_t(\epsilon) = \frac{8}{\epsilon^2}
    \left(1-\sum_{x} \sqrt{
        P_{\vlambda_0 + \vecv \epsilon (t - 1/2)}(x) P_{\vlambda_0 + \vecv \epsilon (t+1/2)}(x)
      }\right).
  \end{equation}
  Let $p_s(x) = P_{\vlambda_0 + \vecv s}(x)$. Then, according to condition 3,
  we can write
  \begin{equation}
    \label{eq:th.bitchifc.proof4}
    p_s(x) = p_0(x) + \dot p_0(x) s + \delta p_s(x), \quad \text{where} \quad
    \dot p_0(x) = \left.\frac{\partial}{\partial s}\right|_{s=0} p_s(x),\quad
    \delta p_s(x) = o(s).
  \end{equation}
  According to condition 4, for any $x$ with $p_0(x) = 0$,
  we have $p_s(x) = o(s^2)$, i.e., $\dot p_0(x) = 0$ and $\delta p_s(x) = o(s^2)$.
  Since $\sum_x p_s(x) = 1$, we have $\sum_x \dot p_0(x) = 0$
  and $\sum_x \delta p_s(x) = 0$.
  By expanding \cref{eq:th.bitchifc.proof3} to $o(1)$ in $\epsilon$
  using $p_0(x)$, $\dot p_0(x)$, and $\delta p_s(x)$, and applying their properties listed above, we see that such an expansion does not depend on $t$.
  Therefore, it remains to show that
  \begin{equation}
    \label{eq:th.bitchifc.proof5}
    f(\epsilon) = \hat{g}(\vlambda_0;\vecv) + o(1).
  \end{equation}
  Since we are only interested in the limit $\epsilon\to 0$, we can assume that $p_s(x) > 0$ for all $x$ with $p_0(x) > 0$ for $\abs{s} \leq \abs{\epsilon}$.
  Let $\vlambda_{+} = \vlambda_0 + \vecv \epsilon/2$,
  $\vlambda_{-} = \vlambda_0 - \vecv \epsilon/2$,
  $p_{+}(x) = p_{1/2}(x)$, $p_{-}(x) = p_{-1/2}(x)$, and $q(x) = (p_+(x) + p_-(x))/2$.
  Then we can write
  \begin{equation}
    \label{eq:th.bitchifc.proof6}
    f(\epsilon) = \frac{8}{\epsilon^2}\left(
    1 - \mathbb{E}_{x\sim q} \sqrt{p_+(x) p_-(x)/q^2(x)}\right) + o(1),
  \end{equation}
  where $o(1)$ absorbed terms with $q(x) = 0$.
  From \cref{aeq:th.bitchifc.1} it follows that for $x$ with $q(x) > 0$ we have
  \begin{equation}
    \label{eq:th.bitchifc.proof7}
    p_+(x) p_-(x)/q^2(x) = (\cosh(\epsilon \vecv \cdot M^*(\vlambda_0, \vecv \epsilon, x)/2))^{-2}.
  \end{equation}
  Substituting \cref{eq:th.bitchifc.proof7} into \cref{eq:th.bitchifc.proof6} we get
  \begin{align}
    f(\epsilon)
    &= \frac{8}{\epsilon^2}
    \mathbb{E}_{x\sim q} \left(
    1-\cosh(\epsilon \vecv \cdot M^*(\vlambda_0, \vecv \epsilon, x)/2)^{-1}\right) + o(1) \notag \\
    &= \mathbb{E}_{x\sim p_0} (\vecv \cdot M^*(\vlambda_0, 0, x))^2 + o(1)
    = \hat{g}(\vlambda_0;\vecv) + o(1).
    \label{eq:th.bitchifc.proof8}
  \end{align}
\end{proof}

\subsection{ClassiFIM BC implementation choices}
\label{ss:classiFIM-BC-implementation}
We summarize the design choices we made in our own implementation of a ClassiFIM BC:
\begin{enumerate}
  \item[(a)] We use a neural network as the trainable function $\vecl = M(\vlambda_0, \vdlambda, x; \vectheta)$ and minimize the cross-entropy error.
    %-------------------
  \item[(b)] We add a small $L_2$ regularization to the cross-entropy cost to avoid overfitting.
    %-------------------
  \item[(c)] We choose a neural network architecture suitable for our problems (e.g., for a periodic lattice, we use a CNN to respect translation symmetry) among other common sense, problem-specific choices.
    %----------------
  \item[(d)] We do not use batch normalization (BN) or other normalization layers.
\end{enumerate}
Aside from outputting a vector $\vecl$ as in (a), the choices in (a) -- (c) are relatively standard for modern state-of-the-art BCs. On the other hand, most modern BCs use BN, so (d) is non-standard. We remark that (d) is known to help train to near-zero loss for classification accuracy. In our work, BN makes it more difficult to regularize the model using $L_2$ regularization, which we find is more suitable for our use case where accurate estimates of log-odds (instead of accurate classifications) are needed.

\subsection{Fine-tuning of algorithm 1}
\label{ass:fine-tune-alg1}

In the actual implementation, we notice that $\mcDtrain$ fits into the GPU memory for the datasets we used in our numerical experiments. To optimize ClassiFIM performance, we use a slightly different algorithm than \cref{alg:generateBC} to generate $\mcDtrainBC$. We first load $\mcDtrain$ (not $\mcDtrainBC$) into the GPU memory. Then, before each epoch, we generate a version of the dataset $\mcDtrainBC$ on the GPU as described in \cref{alg:generateBC2}. In our early experiments, we observed that this optimization improves both computational performance (in samples per second we train on) and generalization (because for each epoch we use a different $\mcDtrainBC$, although it is generated from the same $\mcDtrain$) when compared to generating a single $\mcDtrainBC$ with size $N_{\textnormal{BC}} \simeq 10 \abs{\mcDtrain}$.

We apply the same \cref{alg:generateBC2} to $\mcDtest$ to test the binary classifier on the hold-out set (which is used for reporting the test cross-entropy).

\begin{algorithm}[ht]
  \caption{Generating $\mcDtrainBC$ from $\mcDtrain$ on GPU before each epoch}
  \label{alg:generateBC2}
  \textbf{Input}: dataset $\mcD$\\
  \textbf{Output}: $\mcDBC$
  \begin{algorithmic}[1] %[1] enables line numbers
    \STATE{$\mcD_{+}, \mcD_{-} \leftarrow \textnormal{two random permutations of } \mcD$}
    \STATE{Discard rows $i$ from $\mcD_{\pm}$ with $\vlambda_{+,i} = \vlambda_{-,i}$}
    \STATE{$\mcDBC\leftarrow\varnothing$}
    \STATE{\COMMENT{Execute the following in parallel for each $i$:}}
    \FOR{$i \in \textnormal{range}(\textnormal{len}(\mcD_{+}))$}
    \STATE{$(\vlambda_{+}, x_{+}) \leftarrow \textnormal{row \#i from }\mcD_{+}$}
    \STATE{$(\vlambda_{-}, x_{-}) \leftarrow \textnormal{row \#i from }\mcD_{-}$}
    \STATE{$\lambda_0 \leftarrow (\vlambda_{+} + \vlambda_{-})/2$}
    \STATE{$\vdlambda = \vlambda_{+} - \vlambda_{-}$}
    \STATE{$\mcDBC \leftarrow \mcDBC %
        \cup \{(\vlambda_0, \vdlambda, x_{-}, 0), (\vlambda_0, \vdlambda, x_{+}, 1)\}$} %
    \COMMENT{the last component is the label}
    \ENDFOR
    \STATE \textbf{return} $\mcDBC$
  \end{algorithmic}
\end{algorithm}

\section{Modifications for comparison with prior work}
\label{as:prior-work}

In \cref{sec:comparison} we describe our methodology for comparing ClassiFIM
with W and SPCA using the PeakRMSE metric. In particular, it requires us to
re-implement the W and SPCA methods and to post-process the outputs of all three
methods so that the methods return the guesses $\{x_{s,k}\}_{k=1}^{n_s}$ for each
slice $s$ of the parameter space. We denote the resulting implementations as
mod-W, mod-SPCA, and mod-ClassiFIM. Here the ``mod-'' prefix indicates our modifications
including efficiency improvements and post-processing to obtain the guesses,
``W'' stands for the expected shape of the accuracy plot in
Confusion Scheme of \cite{vanNieuwenburg2016LearningPT},
and ``SPCA'' stands for the Shadow Principal Component Analysis method of \cite{Huang:22}.

We first describe the methods and modifications in detail in
\cref{ass:mod-W,ass:mod-SPCA,ass:mod-ClassiFIM} and then detail in \cref{ass:broken-rules}
the instances where the rules from \cref{sec:comparison} were broken,
explaining why and discussing the anticipated impact.

Some modifications we describe improve the computational efficiency of the method.
We note that these are important to achieve a better PeakRMSE because
better computational efficiency allows us to use a larger network size and number of epochs
in mod-W and avoid subsampling in mod-SPCA within the computational budget,
both of which improve the quality of the resulting predictions.

\subsection{Details of mod-W}
\label{ass:mod-W}
The Confusion Scheme method in Ref.~\cite{vanNieuwenburg2016LearningPT}
produces a ``W''-like shape with the middle peak pointing
at the predicted location of phase transition
(or a ``V''-like shape indicating no phase transition is found).
The method, as described, only applies to systems parameterized
by a single parameter $\lambda$.
Therefore, to apply it to datasets with two parameters,
we apply it separately to each horizontal and vertical slice of that dataset.
Consider, for example, a $64 \times 64$ parameter grid. Then there are $128$ such slices.
For each horizontal slice, $\lambda_1$ is fixed and the method considers $63$ possible values of $\lambda_{0c}$ (possible phase transition locations, midpoints of the original grid).
For each such location, a separate neural network is trained to predict,
given a sample $x$, whether $x$ was sampled from $P(\bullet|\vlambda)$
with $\lambda_0 < \lambda_{0c}$ or $\lambda_0 > \lambda_{0c}$.

\begin{figure*}
  \includegraphics[width=\dimexpr0.99\textwidth\relax]{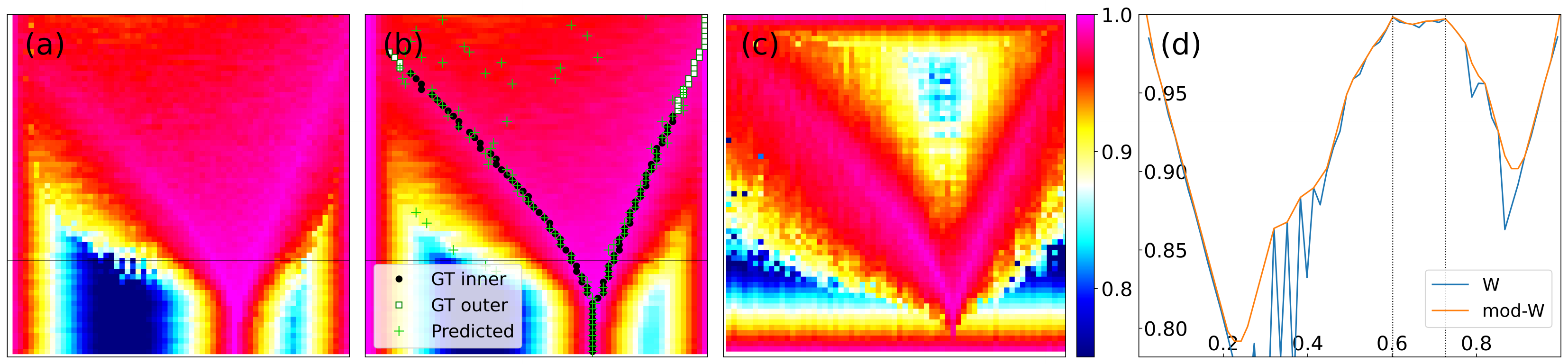}
  \caption{
  Illustration of mod-W.
  \textbf{Panel (a)}: Accuracy plot generated by the W method for horizontal slices,
  assembled into a 2D phase diagram (identical to \cref{fig:isnnn400-1x7}(d)).
  Accuracy values below $75\%$ are truncated to $75\%$ and represented in dark blue.
  \textbf{Panel (b)}: The plot after post-processing, showing the removal of spurious low-accuracy points.
  Ground truth and predicted peaks are overlaid on the 2D diagram.
  Each horizontal slice contains $n_s'$ black circles representing the ``inner'' ground truth peaks, which are the targets for prediction by methods like mod-W.
  Additional $n_s - n_s'$ light squares in each slice represent
  ground truth peaks which are either too shallow or too close to the border,
  and their predictions are not necessary in the PeakRMSE metric
  given by \cref{eq:peakrmse},
  but their count $n_s - n_s'$ represents the additional number of guesses
  each method like mod-W is allowed to make in that slice.
  The $n_s$ green crosses in each slice indicate these guesses made by mod-W.
  For each black circle in each slice, there is a corresponding contribution to the PeakRMSE metric equal to the distance to the nearest green cross in that slice (or the border of the slice if it is closer).
  \textbf{Panel (c)}: Accuracy plot generated by method W for vertical slices within the same computational budget.
  In mod-W, we then post-process it and predict the peaks (not shown, the process is the same as for horizontal slices).
  \textbf{Panel (d)}: A single horizontal slice at $\lambda_1 = 18/64$ (marked on panels (a) and (b) with a horizontal line) with W plot before and after post-processing, i.e., accuracy as a function of $\lambda_0$.
  Vertical dotted lines mark the ground truth peaks.
  For this dataset and slice, the predicted peaks (i.e., local maxima of the mod-W plot)
  align perfectly with the ground truth.
  On panels (a)--(c) the axes are $\lambda_0$ (horizontal) and $\lambda_1$ (vertical) ranging from $0$ to $1$.}
  \label{fig:isnnn400-w-details}
\end{figure*}

One is then expected to plot classification accuracy as a function of $\lambda_{0c}$,
where the peaks indicate the possible locations of phase transitions.
As originally described, the method would require training $128 \cdot 63$
different neural networks for the 2D datasets we use.
This means that, e.g., for the Hubbard12 dataset,
about $0.074$ seconds out of a ten-minute budget is available for training a single neural network,
which is problematic due to the overheads involved.
Therefore, we optimized the training procedure by combining all neural networks
for a given slice into a single larger network: it is mathematically equivalent
to $L-1$ separate networks with identical architectures (but different weights),
but computationally more efficient when training using PyTorch.
Here $L$ is the number of grid points in the slice:
$L=64$ for all 2D datasets we use.
During training, $L-1$ separate networks receive the same samples $x_i$
but different labels corresponding to different values of $\lambda_{0c}$.
We took a neural network architecture similar to the one for ClassiFIM,
removing the parts unnecessary for this simpler task.
Furthermore, we had to reduce the network size and the number
of epochs significantly to fit within the same computational budget.
We used half of $\mcDtrain$ for training and half for evaluation.
This is contrary to other methods that use the whole dataset
for training but is necessary to follow the rules from \cref{sec:comparison}.
Indeed, in the W method, predicted phase transition locations are extracted from
the plot of validation accuracy as a function of $\lambda_{0c}$,
which requires a validation set separate from the training set,
but the rules require that only $\mcDtrain$ is used for making such predictions,
thus requiring us to split $\mcDtrain$.
The resulting accuracy plots (stitched together into one 2D phase diagram,
with purple indicating perfect accuracy and blue indicating $75\%$ accuracy)
are displayed in \cref{fig:isnnn400-w-details}(a).

In order to post-process the predicted accuracy plots,
observe that if $\textrm{NN}_j$ is trained to predict whether
$\lambda_0 < \lambda_{0cj}$ for $j = 1,2$ and the resulting accuracies
are $a_j$, then one would expect
\begin{equation}
a_2 \geq a_1 - \abs{\lambda_{0c1} - \lambda_{0c2}}
\label{eq:w-accuracy-bound}
\end{equation}
(assuming the uniform distribution of the values of $\lambda_0$ in the validation set).
Yet, in practice, this constraint is sometimes violated because of imperfect training
(as can be seen from the blue points in \cref{fig:isnnn400-1x7}(d) of the main text).
This motivates the first two steps of the postprocessing we use for W:
  (1) pad the list of accuracies by $1.0$ on both sides;
  (2) reset the accuracy values to the maximum of the r.h.s. of \cref{eq:w-accuracy-bound}.
The final post-processing step is to further reduce the spurious points with low accuracy:
if the arithmetic mean of accuracies of two neighboring points is greater
than the accuracy at the current point,
we increase the current estimated accuracy to that value.

Finally, per the methodology of PeakRMSE \citeCompanionPaper[Appendix L]{}, for each slice,
we are given a number of peaks (i.e., the maximal number of guesses) $n_s$
and need to return their predicted locations. We do this by
extracting the highest prominence peaks using \verb!scipy.signal.find_peaks!.
When the number of peaks thus extracted is less than $n_s$,
we add additional guesses
to split the largest gaps as evenly as possible: e.g., if we are missing one peak,
we add a guess at the midpoint of the largest gap.
This process is illustrated in \cref{fig:isnnn400-1x7}.

\subsection{Details of mod-SPCA}
\label{ass:mod-SPCA}
In all physics datasets, bitstrings $x_i\in \mcDtrain$ are used as samples.
I.e., it is natural to use individual bits of $x_i$ as features.
This is in contrast to, e.g., MNIST-CNN, where the features of $x_i$ are naturally interpreted as real numbers and not individual bits.
\citeCompanionPaper[App J]{} described how to apply SPCA to such bitstring datasets.
Specifically, we need to apply kernel PCA to the kernel given by
\begin{equation}
  \label{eq:spca-bs-kernel}
  K(\vlambda, \vlambda') = \exp\Biggl(
    \frac{\tau}{T_{\vlambda} T_{\vlambda'}}
    \sum_{t=1}^{T_{\vlambda}} \sum_{t'=1}^{T_{\vlambda'}}\exp\left(
      \frac{\gamma}{n} \sum_{i=1}^{n} I\left(
        x_{\vlambda,i}^{(t)} = x_{\vlambda',i}^{(t')}
      \right)\right)\Biggr).
\end{equation}
Here $\vlambda$, $\vlambda'$ are two points in the parameter space,
$T_{\vlambda}$ is the number of pairs $(\vlambda, x) \in \mcDtrain$
with a given $\vlambda$, $\{x_{\vlambda}^{(t)}\}_{t=1}^{T_{\vlambda}}$
are the corresponding samples, $x_{\vlambda,i}^{(t)}$ is the $i$-th
bit of $x_{\vlambda}^{(t)}$ and $I$ is the indicator function.
The kernel is parameterized by two hyperparameters: $\tau, \gamma > 0$.

We pick $\tau = 1$ and $\gamma = 1$ as in \cite{Huang:22}.
We note that \citeCompanionPaper[App J.3]{} suggests that
an alternative interpretation of the original SPCA method
when applied to bitstring datasets is to pick $\tau = e^{-1}$ and $\gamma = 1$
in \cref{eq:spca-bs-kernel}. However, we did not test these because
we are not aware of any method
to test the performance of different choices of hyperparameters
in SPCA without a comparison to the ground truth, and such a
comparison would violate the rules set in \cref{sec:comparison}.
This is in contrast to W and ClassiFIM, where we can split
a validation set out of $\mcDtrain$ and compute the training accuracy
and the training loss respectively to evaluate the choice of hyperparameters.

Our first modification is efficiency. The most computationally expensive part
of SPCA is the evaluation of the kernel matrix using \cref{eq:spca-bs-kernel},
which requires $O(M^2 T^2 n)$ operations,
where $M = \abs{\mcM'}$ is the number of grid points in the parameter space,
$T\simeq 135$ is the maximal number of samples for a single grid point,
and $n$ is the number of bits in a sample.
A naive implementation would have five nested loops.
However, the inner loop
in \cref{eq:spca-bs-kernel} can be computed efficiently using
the bitwise exclusive or ($\oplus$, XOR)
and the \verb!POPCNT! instruction available on modern CPUs.
The latter counts the number of bits set to $1$ in a $64$-bit integer,
typically with a throughput of one instruction per cycle (and latency of $\sim 3$ cycles),
which is much faster than a loop.
This assumes that the bitstrings are stored as integers.
We use 64-bit integers in our implementation and note that
for larger $n$ we could also take advantage of one of SIMD instruction
sets but this was not done in our implementation.
For $n\leq 64$ we have
\begin{equation}
  \sum_{i=1}^{n} I\left(x_{\vlambda,i}^{(t)} = x_{\vlambda',i}^{(t')}\right)
  = n - \textnormal{POPCNT}(x_{\vlambda}^{(t)} \oplus x_{\vlambda'}^{(t')}).
\end{equation}
For larger $n$, we used $\ceil{n/64}$ such \verb!POPCNT! operations.

The result of the SPCA is illustrated in \cref{fig:isnnn400-1x7}(e) of the main text,
where the red, green, and blue color channels correspond to the top three PCA components scaled to $[0, 1]$.
To extract the peak locations from a horizontal slice, we first apply smoothing to the top two PCA components (along $\lambda_0$),
which is identical to the ClassiFIM post-processing explained below. Then, we create a list of features by appending a monotonically increasing (with $\lambda_0$) feature to the list of PCA components and run \verb!sklearn.cluster.KMeans!, and adjust the clusters so that they are contiguous in $\lambda_0$. Cluster boundaries are then the returned peak locations.
While technically the total time often exceeds the time limit,
this is because we did not attempt to further parallelize
the shadow matrix computation or port it to GPU. For example,
running the computation for ten Hubbard12 datasets in parallel
on a single i7-6850K CPU results in a total computation time of about $30$ minutes,
i.e., three minutes per dataset (less than the $10$ minutes budget).

\subsection{Details of mod-ClassiFIM}
\label{ass:mod-ClassiFIM}
\begin{figure*}
  \includegraphics[width=\dimexpr0.99\textwidth\relax]{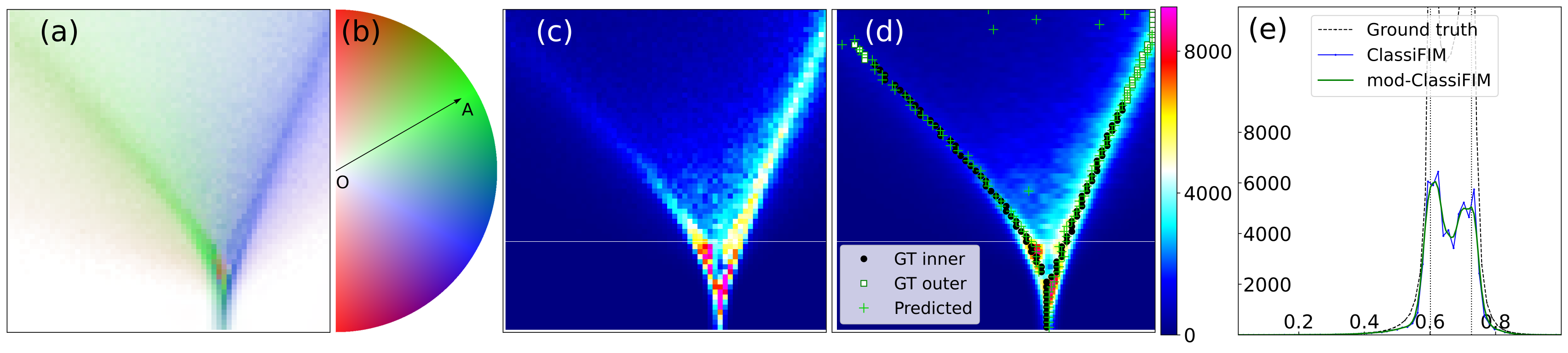}
  \caption{%
  Illustration of mod-ClassiFIM.
  \textbf{Panel (a)}: The ClassiFIM prediction from \cref{fig:isnnn400-1x7}(f).
  Darker colors denote higher values of $\Tr(\hat{g}(\vlambda))$.
  To compute the intensity $I_c$ of color channels $c \in \{\mathrm{red}, \mathrm{green}, \mathrm{blue}\}$, we selected three unit vectors $\vecv_c$, each separated by $2\pi/3$, and computed $I_c = 1 - \max(1, \sqrt{\hat{g}(\vlambda; \vecv_c)} / C)$ where $C = 140$ serves as a normalization constant and $\hat{g}(\vlambda; \vecv) = \sum_{\mu\nu} \hat{g}_{\mu\nu}(\vlambda)v_{\mu} v_{\nu}$ is the squared length of vector $\vecv$ as per the metric $\hat{g}(\vlambda)$.
  This color scheme was also used in panels (f) and (g) of \cref{fig:isnnn400-1x7}.
  \textbf{Panel (b)}: Semidisk illustrating the color scheme. For each point A in the semidisk, we define a vector $\vecv = \protect\overrightarrow{OA}$ (an example of such a vector is shown). Then the color of such a point A represents $g$ defined using $g_{\mu\nu} = C^2 v_{\mu} v_{\nu}$.
  For instance, the color of point O is white because it corresponds to $\vecv = 0$ and, hence, $g=0$.
  Such a semicircle only shows the colors corresponding to rank-1 tensors $g$: for example, the color of $g = C^2 I$ would be black (not shown).
  \textbf{Panel (c)}: $\hat{g}_{00}$, the metric component to be postprocessed in the horizontal slices.
  \textbf{Panel (d)}: Post-processed $\hat{g}_{00}$. Ground truth and predicted peaks are overlaid on the 2D diagram similarly to panel (b) of \cref{fig:isnnn400-w-details}.
  \textbf{Panel (e)}: A single horizontal slice at $\lambda_1 = 18/64$ (indicated in panels (b) and (c) using a horizontal line) showing the raw and post-processed $\hat{g}_{00}$ values alongside the ground truth.
  Vertical dotted lines denote the positions of ground truth peaks. Nearby local maxima of the post-processed $\hat{g}_{00}$ are the predicted peak locations.}
  \label{fig:isnnn400-cf-details}
\end{figure*}
To predict peaks for each horizontal slice with ClassiFIM, we take the predicted $\hat{g}_{00}$ and apply the smoothing along the $\lambda_0$ direction by convolving $\hat{g}_{00}$ with a Gaussian kernel. We use two versions of the kernel shifted by half of the lattice spacing to double the resolution of the resulting function. Similarly to mod-W, we extract the highest value peaks using \verb!scipy.signal.find_peaks!.
This is illustrated in \cref{fig:isnnn400-cf-details}. Vertical slices are processed analogously.

\subsection{Potential data snooping in evaluating PeakRMSE metric}
\label{ass:broken-rules}
Given the limited number of datasets and low statistical power of the PeakRMSE metric
(as demonstrated in \cref{ss:comparison-of-metrics}), it is important to avoid
introducing bias into the result of the comparison of the mod-W, mod-SPCA, and mod-ClassiFIM methods.
This is why \cref{sec:comparison} introduced a set of four rules one might follow
to make the comparison fair.

Strict adherence to these rules is nontrivial,
and below we point out when a rule was broken, to what extent, and why.
The whole process involved trial and error, violating rule (ii):
the PeakRMSE metric was originally designed only for FIL24 and,
more specifically, only for horizontal slices.
In its original formulation, it suffered from issues listed in
\citeCompanionPaper[App. L.1]{} resulting in the value of the metric
being dominated by a single slice where a method's mistake could be explained
by a small misjudgment in the location or height of the peak.
To avoid these issues, the definition of PeakRMSE was adjusted in multiple
iterations, each of which occurred only after we observed the results.
In its original form, it also had a version that only considered slices
with a single phase transition. This was because
Ref.~\cite{vanNieuwenburg2016LearningPT} only considered a case
of a single phase transition. However, we observed that
the performance of W was not significantly affected by whether the slice
had one or more phase transitions and decided to use all available
slices, which, in theory, should improve the statistical power of the metric.

The specific modifications made to each method were made only after
observing the outputs of the methods on our datasets. For example,
we observed that the actual accuracy curve returned by the W method
often did not reach values close to 1 at the locations of phase transitions,
motivating us to use the highest prominence peaks. However, such highest prominence
was affected by a few points with exceptionally low accuracy, which
motivated the slope-capping post-processing step. The original clustering
for SPCA operated on the top ten PCA components, but we observed that
this resulted in worse clusters than what one can guess by eye by
looking at the 2D phase diagram with the top three PCA components,
used as three color channels, motivating us to reduce the number of PCA components to two.

The design of the original NN architectures involved some trial and error.
Most of such trials were done on lower quality datasets not
used in the final comparison. Yet, some choices were still made when
training on the current datasets, but without looking at the ground truth.
In such cases, the choice was made based on the training loss (for ClassiFIM)
or accuracy (for W) on the training or validation set (which was split out of $\mcDtrain$).
However, such decisions were made once and not separately for each seed.
Additionally, the time spent on such tuning was not included in the
computational budget.

\section{Comparison methods}
\label{as:comparison}

In \cref{tab:qpt-results}, we show the performance of ClassiFIM on physics datasets.
The rows ``const'' and ``best'' are added as a comparison.
In ClassiFIM, a binary classification model is used to provide the $\hat{g}$ estimates.
On the other hand, for the comparison rows, the binary classification predictions are not related to the predictions of $g$.
Below, we provide the definitions of these comparison rows.

\subsection{``Best''}
For binary classification, ``best'' produces the best possible
probability estimates equal to the ground truth \textit{a posteriori} probabilities
$P(y=1| x, \vlambda_0, \vdlambda) = P_{\vlambda_{+}}(x) / (P_{\vlambda_{+}}(x) + P_{\vlambda_{-}}(x))$.
The standard deviation is non-zero because different
runs had different test sets.

The ``best possible'' classical fidelity susceptibility predictions are
equal to the ground truth $g_{\mu\nu}(\vlambda)$ (contrary to ClassiFIM,
they are not generated for the grid points in $\mcM'$ but for the shifted grid $\mcM''$).

\subsection{``Const''}

For binary classification, ``const'' generates the best possible probability
independent of $x$. This probability is equal to $0.5$ since $x$
has the same chance of being sampled from $P_{\vlambda_{+}}$ as from $P_{\vlambda_{-}}$.

The ``const'' classical fidelity susceptibility predictions are
$\hat{g}(\vlambda) = \alpha (d\lambda_0^2 + d\lambda_1^2)$
where $\alpha$ is a positive constant. Note that $\distMSEPS$ and $\distRE$
are scale-invariant, hence their values are not affected by the choice of $\alpha$.
One could pick the value of $\alpha$ to minimize $\distMSE$, in which
case $\distMSE$ would be equal to $\distMSEPS$.

Note further that ``best'' is not really a ``method'' one could apply to solve the
FIM-estimation task, as in the setting of that
task one would not know the ground truth $g$ or the probabilities
used for the ``best'' row. One would not know the perfect
calibration constant $\alpha$ used in the ``const'' method for the $\distMSE$ metric either.
On the other hand, achieving a ``const'' $\distRE$ and $\distMSEPS$
score is easy within the rules
of the task because these two metrics are scale-invariant;
hence one can pick any positive $\alpha$.

\subsection{Naive approach to estimating the FIM}
\label{ass:naive-approach}
Consider, for example, that samples $x$ are bitstrings of length $n$
(as is the case in all physics datasets we use in this work),
and let $\mcS = \{0, 1\}^n$ be the set of all possible samples.
Suppose we have just one parameter $\lambda$ and the dataset contains
$N$ samples for $\lambda = \lambda_0 - \delta/2$ and $N$ samples for
$\lambda = \lambda_0 + \delta/2$.
Then we could try to estimate $g(\lambda_0)$ as:
\begin{equation}
  \hat{g}(\lambda_0)
  = \frac{8}{\delta^2}\left(1 - \hat F_c\right)
  = \frac{8}{\delta^2}\left(1 - \sum_{x \in \mcS}
  \frac{\sqrt{N_{x, {-}} N_{x, {+}}}}{N}\right).
\end{equation}
Here $N_{x, {\pm}}$ are the numbers of samples equal to $x$
for $\lambda = \lambda_0 \pm \delta/2$.
That is, one would use $N_{x, {\pm}}/N$ as
estimates of the probabilities of $x$ at $\lambda_0 \pm \delta/2$.

To see a problem with this estimate, consider an example where
all probability distributions $P_{\lambda}$ are identical, i.e.,
$g(\lambda_0) = 0$ and $F_c(\lambda_0 - \delta/2, \lambda_0 + \delta/2)=1$.
Since the number of possible bitstrings
is exponential in the length of the bitstrings (e.g., $2^{400}$ for IsNNN400),
one might expect that typically
$\mathbb{E}(N_{x,-}) = \mathbb{E}(N_{x,+}) = p_{x} N \ll 1$ for all $x$,
where $p_{x} = P_{\lambda_{+}}(x) = P_{\lambda_{-}}(x)$.
In this case, however, we have
\begin{equation}
  \mathbb{E}\left(\frac{\sqrt{N_{x,-}N_{x,+}}}{N}\right)
  = \frac{1}{N}\left(\mathbb{E}\left(\sqrt{N_{x,-}}\right)\right)^2
  = p_x^2 N (1 + O(p_{x}N)).
\end{equation}
i.e., if $\varepsilon = \max_x p_{x} N \ll 1$, then
\begin{equation}
  \mathbb{E}(\hat F_c) < \varepsilon + O(\varepsilon^2).
\end{equation}
Thus, we might expect such an estimate $\hat{g}(\lambda_0)$ to be close to $8/\delta^2$
and far from the true value $0$. For this reason, we did not use this naive approach in our work.

\section{Limitations and future work}
\label{as:limitations}
\Cref{s:limits-and-future} outlined the limitations of this work and
suggested directions for future work. Here we discuss these topics in more detail.

\subsection{The FIM-estimation task is not always solvable}
\label{ass:lim-task}
Here we provide an example of an FIM-estimation task which,
one might expect, would not be solvable by ClassiFIM or
any other efficient method.

Consider a cryptographic hash function $f$ and
a SM where the only parameter is $\lambda \in [-1, 1]$
and the corresponding probability distributions $P_{\lambda}$ are over
bitstrings $y = f(x)$, where $x$ is distributed uniformly among bitstrings
with $x_l = v(\lambda)$ for some fixed index $l$ and some fixed
function $v$ with values in $\{0, 1\}$ (that is, one bit of $x$ is set to a fixed value dependent on $\lambda$,
while all other bits are uniformly random).
The samples are provided as part of $\mcDtrain$, and the full description
of the generation procedure including the definition of function $f$
but excluding the choice of $v$ and $l$ are given as part of the textual description.

If an algorithm can estimate the FIM
efficiently, it could distinguish between the constant function $v$ and
a function experiencing a jump at $\lambda = 0$. That is, the algorithm
would efficiently predict a property of the pre-images of the hash function $f$,
which, one might expect, is not possible (more precisely, this is not possible under the random oracle model of cryptographic hash functions).

\subsection{Use of local features}
\label{as:lim-local}
A typical architecture used for ClassiFIM NNs within this work is illustrated in
\cref{fig:nn-architecture}. Samples $x_i$ are first processed by a CNN. The output
of the CNN is the result of (possibly multiple) local convolutions and non-linearities,
which can result in output values depending only on parts of $x_i$ corresponding to
a small neighborhood of the lattice site associated with the output value. Consequently,
these features are averaged over all lattice sites and fed into a fully connected layer.
In other words, the output of the NN can be described as
\begin{equation}
  \label{eq:local-features}
  g\left(\left\{\frac{1}{N} \sum_{j} f_{\alpha, j}(x_i)\right\}_{\alpha}\right),
\end{equation}
where $N$ is the number of sites, $j$ indexes the sites, and $f_{\alpha, j}(x_i)$ is
the local feature $\#\alpha$ at site $j$. This precludes the NN from computing
features like the parity of bits in $x_i$ along a (long non-local) loop in the lattice,
since $f_{\alpha, j}(x_i)$ are local and used by a permutation-invariant
(with respect to $j$) expression \cref{eq:local-features}.

If two distributions are only globally distinguishable,
as in the case of some topological quantum phase transitions
(TQPTs, see, e.g., \cite{hamma2008entanglement,Wen:2017aa}),
it is possible that our method with such NN architectures would fail
to estimate the underlying FIM, though, as we show in \citeCompanionPaper[Fig. 6, 10]{},
ClassiFIM produces accurate FIM phase diagrams for the XXZ and Kitaev models,
both of which are known to exhibit TQPTs~\cite{Kitaev:97,Elben}.

Neverthless, this is not a limitation of ClassiFIM itself because ClassiFIM
does not require any particular use of NN architecture, as we discuss below.

\subsection{Other NN architectures}
\label{as:lim-architectures}
The ClassiFIM method is explained in \cref{sec:classifim} (steps 1--3).
None of these steps require any particular NN architecture. Rather,
some BC model is needed as explained in step 2.
Within this work, we chose to use a simple architecture as shown in \cref{fig:nn-architecture} as such BC model. Future work may find that some other NN architecture would achieve better performance if used within ClassiFIM.
For example, one could try a positional encoding for the $\vlambda$ input, possibly with an alternative way of encoding the set $\{\vlambda_{-}, \vlambda_{+}\}$: e.g., instead or in addition to supplying $\vdlambda\vdlambda^T$, one may supply this
set via a (two-element) attention layer. One may also try non-neural network models
as binary classifiers within ClassiFIM.

We did not use any normalization layers in our NN architectures because
we did not obtain good performance using them in our early experiments
(on synthetic datasets not reported in this work).
We conjecture that, in the context of classification tasks,
such layers are significantly more effective in the modern regime
when one aims to maximize the hold-out accuracy,
than for problems where ground truth probabilities are far from $0$ or $1$ and those probabilities are desired (e.g., when the goal is to minimize the hold-out cross-entropy error).
A possible cause for this could be that normalization layers change the role of
weight decay in the previous layers, making the regularization harder.
Future work could investigate this hypothesis, provide a theoretical explanation for this observation,
and possibly use such an explanation to design better NN architectures for the latter type of problems.

\subsection{Other tasks}
\label{as:lim-other}
Future work may investigate whether ClassiFIM can achieve good performance on
other problems (possibly where the formal justification in \cref{th:bitchifc}
does not apply).

In \cref{ss:mnist-cnn-results}, we considered the detection
of qualitative changes in the results of CNN training as hyperparameters are varied.
Another potentially more interesting application of ClassiFIM arises when a
Boltzmann machine is used instead of a CNN: one may look for a qualitative change
in the distribution of states of a layer of the Boltzmann machine as some
parameters are varied. Such a problem would be a natural application of the ClassiFIM
because such distributions are naturally interpreted as a statistical manifold.

One could also try to use ClassiFIM for change-point detection
and compare it with other methods (e.g., \cite{aminikhanghahi2017survey,truong2020selective}).

%% START: embedded bu2.bib

%% END: embedded bu2.bib
\end{bibunit}

\begin{thebibliography}{22}
\providecommand{\natexlab}[1]{#1}
\providecommand{\url}[1]{\texttt{#1}}
\expandafter\ifx\csname urlstyle\endcsname\relax
  \providecommand{\doi}[1]{doi: #1}\else
  \providecommand{\doi}{doi: \begingroup \urlstyle{rm}\Url}\fi

\bibitem[Amari(1998)]{amari1998natural}
S.-I. Amari.
\newblock Natural gradient works efficiently in learning.
\newblock \emph{Neural computation}, 10\penalty0 (2):\penalty0 251--276, 1998.
\newblock \doi{https://doi.org/10.1162/089976698300017746}.

\bibitem[Amari(2016)]{amari2016information}
S.-i. Amari.
\newblock \emph{Information geometry and its applications}, volume 194.
\newblock Springer, Japan, 2016.
\newblock \doi{10.1007/978-4-431-55978-8}.
\newblock URL \url{https://link.springer.com/book/10.1007/978-4-431-55978-8}.

\bibitem[Arnold et~al.(2023)Arnold, L{\"o}rch, Holtorf, and
  Sch{\"a}fer]{arnold2023machine}
J.~Arnold, N.~L{\"o}rch, F.~Holtorf, and F.~Sch{\"a}fer.
\newblock Machine learning phase transitions: Connections to the fisher
  information.
\newblock \emph{arXiv preprint arXiv:2311.10710}, 2023.

\bibitem[Baldassi et~al.(2022)Baldassi, Lauditi, Malatesta, Pacelli, Perugini,
  and Zecchina]{baldassi2022learning}
C.~Baldassi, C.~Lauditi, E.~M. Malatesta, R.~Pacelli, G.~Perugini, and
  R.~Zecchina.
\newblock Learning through atypical phase transitions in overparameterized
  neural networks.
\newblock \emph{Physical Review E}, 106\penalty0 (1):\penalty0 014116, 2022.

\bibitem[Chencov(2000)]{chencov1972statistical}
N.~N. Chencov.
\newblock \emph{Statistical decision rules and optimal inference}.
\newblock Number~53. American Mathematical Soc. (originally published in
  Russian, Nauka, 1972), 2000.
\newblock URL \url{https://bookstore.ams.org/mmono-53}.

\bibitem[Duy et~al.(2022)Duy, Nguyen, Nguyen, Trung, and
  Abed-Meraim]{duy2022fisher}
T.~T. Duy, L.~V. Nguyen, V.-D. Nguyen, N.~L. Trung, and K.~Abed-Meraim.
\newblock Fisher information neural estimation.
\newblock In \emph{2022 30th European Signal Processing Conference (EUSIPCO)},
  pages 2111--2115. IEEE, 2022.

\bibitem[Huang et~al.(2022)Huang, Kueng, Torlai, Albert, and
  Preskill]{Huang:22}
H.-Y. Huang, R.~Kueng, G.~Torlai, V.~V. Albert, and J.~Preskill.
\newblock Provably efficient machine learning for quantum many-body problems.
\newblock \emph{Science}, 377\penalty0 (6613):\penalty0 eabk3333, 2022.
\newblock \doi{10.1126/science.abk3333}.
\newblock URL \url{https://www.science.org/doi/abs/10.1126/science.abk3333}.

\bibitem[Ishida et~al.(2020)Ishida, Yamane, Sakai, Niu, and
  Sugiyama]{ishida2020we}
T.~Ishida, I.~Yamane, T.~Sakai, G.~Niu, and M.~Sugiyama.
\newblock Do we need zero training loss after achieving zero training error?
\newblock In \emph{Proceedings of the 37th International Conference on Machine
  Learning}, ICML'20. JMLR.org, 2020.
\newblock URL \url{https://dl.acm.org/doi/abs/10.5555/3524938.3525366}.

\bibitem[Karakida et~al.(2019)Karakida, Akaho, and
  Amari]{karakida2019universal}
R.~Karakida, S.~Akaho, and S.-i. Amari.
\newblock Universal statistics of fisher information in deep neural networks:
  Mean field approach.
\newblock In \emph{The 22nd International Conference on Artificial Intelligence
  and Statistics}, pages 1032--1041. PMLR, 2019.
\newblock URL
  \url{http://proceedings.mlr.press/v89/karakida19a/karakida19a.pdf}.

\bibitem[Kasatkin(2024)]{kasatkin2024classifim-code}
V.~Kasatkin.
\newblock Classifim, 2024.
\newblock URL \url{https://github.com/USCqserver/classifim}.

\bibitem[Kasatkin et~al.(2024)Kasatkin, Mozgunov, Ezzell, and
  Lidar]{kasatkin2024classifim-physics}
V.~Kasatkin, E.~Mozgunov, N.~Ezzell, and D.~Lidar.
\newblock {Detecting Quantum and Classical Phase Transitions via Unsupervised
  Machine Learning of the Fisher Information Metric}, 2024.

\bibitem[Kingma and Ba(2014)]{kingma2014adam}
D.~P. Kingma and J.~Ba.
\newblock Adam: A method for stochastic optimization.
\newblock \emph{arXiv preprint arXiv:1412.6980}, 2014.
\newblock URL \url{https://arxiv.org/abs/1412.6980}.

\bibitem[Kirkpatrick et~al.(2017)Kirkpatrick, Pascanu, Rabinowitz, Veness,
  Desjardins, Rusu, Milan, Quan, Ramalho, Grabska-Barwinska,
  et~al.]{kirkpatrick2017overcoming}
J.~Kirkpatrick, R.~Pascanu, N.~Rabinowitz, J.~Veness, G.~Desjardins, A.~A.
  Rusu, K.~Milan, J.~Quan, T.~Ramalho, A.~Grabska-Barwinska, et~al.
\newblock Overcoming catastrophic forgetting in neural networks.
\newblock \emph{Proceedings of the national academy of sciences}, 114\penalty0
  (13):\penalty0 3521--3526, 2017.
\newblock \doi{https://doi.org/10.1073/pnas.1611835114}.

\bibitem[LeCun et~al.(1998)LeCun, Cortes, and Burges]{lecun1998mnist}
Y.~LeCun, C.~Cortes, and C.~J. Burges.
\newblock The mnist database of handwritten digits, 1998.
\newblock URL \url{http://yann.lecun.com/exdb/mnist/}.

\bibitem[Martens(2020)]{martens2020new}
J.~Martens.
\newblock New insights and perspectives on the natural gradient method.
\newblock \emph{Journal of Machine Learning Research}, 21\penalty0
  (146):\penalty0 1--76, 2020.
\newblock URL \url{http://jmlr.org/papers/v21/17-678.html}.

\bibitem[Oikarinen(2021)]{tuomaso2021mnist}
T.~Oikarinen.
\newblock Training a nn to 99\% accuracy on mnist in 0.76 seconds:
  https://github.com/tuomaso/train\_mnist\_fast, 2021.
\newblock URL \url{https://github.com/tuomaso/train\_mnist\_fast}.

\bibitem[Schervish(1995)]{Schervish:1995aa}
M.~J. Schervish.
\newblock \emph{Theory of Statistics}.
\newblock Springer series in statistics. Springer New York New York, NY, New
  York, NY, 1995.
\newblock ISBN 9781461242505; 1461242509.
\newblock \doi{10.1007/978-1-4612-4250-5}.

\bibitem[Smith and Topin(2019)]{smith2019super}
L.~N. Smith and N.~Topin.
\newblock Super-convergence: Very fast training of neural networks using large
  learning rates.
\newblock In \emph{Artificial intelligence and machine learning for
  multi-domain operations applications}, volume 11006, pages 369--386. SPIE,
  2019.
\newblock \doi{https://doi.org/10.1117/12.2520589}.

\bibitem[Takahashi and Hukushima(2019)]{takahashi2019phase}
J.~Takahashi and K.~Hukushima.
\newblock Phase transitions in quantum annealing of an np-hard problem detected
  by fidelity susceptibility.
\newblock \emph{Journal of Statistical Mechanics: Theory and Experiment},
  2019\penalty0 (4):\penalty0 043102, 2019.

\bibitem[Truong et~al.(2020)Truong, Oudre, and Vayatis]{truong2020selective}
C.~Truong, L.~Oudre, and N.~Vayatis.
\newblock Selective review of offline change point detection methods.
\newblock \emph{Signal Processing}, 167:\penalty0 107299, 2020.
\newblock \doi{https://doi.org/10.1016/j.sigpro.2019.107299}.

\bibitem[van Nieuwenburg et~al.(2016)van Nieuwenburg, Liu, and
  Huber]{vanNieuwenburg2016LearningPT}
E.~P.~L. van Nieuwenburg, Y.-H. Liu, and S.~D. Huber.
\newblock Learning phase transitions by confusion.
\newblock \emph{Nature Physics}, 13:\penalty0 435 -- 439, 2016.
\newblock \doi{10.1038/nphys4037}.
\newblock URL \url{https://www.nature.com/articles/nphys4037}.

\bibitem[Zhang et~al.(2021)Zhang, Bengio, Hardt, Recht, and
  Vinyals]{zhang2021understanding}
C.~Zhang, S.~Bengio, M.~Hardt, B.~Recht, and O.~Vinyals.
\newblock Understanding deep learning (still) requires rethinking
  generalization.
\newblock \emph{Communications of the ACM}, 64\penalty0 (3):\penalty0 107--115,
  2021.
\newblock \doi{https://doi.org/10.1145/3446776}.

\end{thebibliography}

\begin{thebibliography}{13}
\providecommand{\natexlab}[1]{#1}
\providecommand{\url}[1]{\texttt{#1}}
\expandafter\ifx\csname urlstyle\endcsname\relax
  \providecommand{\doi}[1]{doi: #1}\else
  \providecommand{\doi}{doi: \begingroup \urlstyle{rm}\Url}\fi

\bibitem[Amari(2016)]{amari2016information}
S.-i. Amari.
\newblock \emph{Information geometry and its applications}, volume 194.
\newblock Springer, Japan, 2016.
\newblock \doi{10.1007/978-4-431-55978-8}.
\newblock URL \url{https://link.springer.com/book/10.1007/978-4-431-55978-8}.

\bibitem[Aminikhanghahi and Cook(2017)]{aminikhanghahi2017survey}
S.~Aminikhanghahi and D.~J. Cook.
\newblock A survey of methods for time series change point detection.
\newblock \emph{Knowledge and information systems}, 51\penalty0 (2):\penalty0
  339--367, 2017.
\newblock \doi{10.1007/s10115-016-0987-z}.

\bibitem[Elben et~al.(2020)Elben, Yu, Zhu, Hafezi, Pollmann, Zoller, and
  Vermersch]{Elben}
A.~Elben, J.~Yu, G.~Zhu, M.~Hafezi, F.~Pollmann, P.~Zoller, and B.~Vermersch.
\newblock Many-body topological invariants from randomized measurements in
  synthetic quantum matter.
\newblock \emph{Science Advances}, 6\penalty0 (15):\penalty0 eaaz3666, 2020.
\newblock \doi{10.1126/sciadv.aaz3666}.

\bibitem[Hamma et~al.(2008)Hamma, Zhang, Haas, and
  Lidar]{hamma2008entanglement}
A.~Hamma, W.~Zhang, S.~Haas, and D.~Lidar.
\newblock Entanglement, fidelity, and topological entropy in a quantum phase
  transition to topological order.
\newblock \emph{Physical Review B}, 77\penalty0 (15):\penalty0 155111, 2008.
\newblock \doi{10.1103/PhysRevB.77.155111}.
\newblock URL \url{https://link.aps.org/doi/10.1103/PhysRevB.77.155111}.

\bibitem[Huang et~al.(2022)Huang, Kueng, Torlai, Albert, and
  Preskill]{Huang:22}
H.-Y. Huang, R.~Kueng, G.~Torlai, V.~V. Albert, and J.~Preskill.
\newblock Provably efficient machine learning for quantum many-body problems.
\newblock \emph{Science}, 377\penalty0 (6613):\penalty0 eabk3333, 2022.
\newblock \doi{10.1126/science.abk3333}.
\newblock URL \url{https://www.science.org/doi/abs/10.1126/science.abk3333}.

\bibitem[Kasatkin(2024{\natexlab{a}})]{kasatkin2024classifim-code}
V.~Kasatkin.
\newblock Classifim, 2024{\natexlab{a}}.
\newblock URL \url{https://github.com/USCqserver/classifim}.

\bibitem[Kasatkin(2024{\natexlab{b}})]{public-datasets}
V.~Kasatkin.
\newblock {Public datasets}, 2024{\natexlab{b}}.
\newblock URL \url{https://huggingface.co/datasets/fiktor/FIM-Estimation}.

\bibitem[Kasatkin et~al.(2024)Kasatkin, Mozgunov, Ezzell, and
  Lidar]{kasatkin2024classifim-physics}
V.~Kasatkin, E.~Mozgunov, N.~Ezzell, and D.~Lidar.
\newblock {Detecting Quantum and Classical Phase Transitions via Unsupervised
  Machine Learning of the Fisher Information Metric}, 2024.

\bibitem[Kitaev(2003)]{Kitaev:97}
A.~Y. Kitaev.
\newblock Fault-tolerant quantum computation by anyons.
\newblock \emph{Annals of Physics}, 303\penalty0 (1):\penalty0 2--30, 2003.
\newblock \doi{http://dx.doi.org/10.1016/S0003-4916(02)00018-0}.
\newblock URL
  \url{http://www.sciencedirect.com/science/article/pii/S0003491602000180}.

\bibitem[Truong et~al.(2020)Truong, Oudre, and Vayatis]{truong2020selective}
C.~Truong, L.~Oudre, and N.~Vayatis.
\newblock Selective review of offline change point detection methods.
\newblock \emph{Signal Processing}, 167:\penalty0 107299, 2020.
\newblock \doi{https://doi.org/10.1016/j.sigpro.2019.107299}.

\bibitem[van Nieuwenburg et~al.(2016)van Nieuwenburg, Liu, and
  Huber]{vanNieuwenburg2016LearningPT}
E.~P.~L. van Nieuwenburg, Y.-H. Liu, and S.~D. Huber.
\newblock Learning phase transitions by confusion.
\newblock \emph{Nature Physics}, 13:\penalty0 435 -- 439, 2016.
\newblock \doi{10.1038/nphys4037}.
\newblock URL \url{https://www.nature.com/articles/nphys4037}.

\bibitem[Wen(2017)]{Wen:2017aa}
X.-G. Wen.
\newblock Colloquium: Zoo of quantum-topological phases of matter.
\newblock \emph{Reviews of Modern Physics}, 89\penalty0 (4):\penalty0 041004--,
  12 2017.
\newblock \doi{10.1103/RevModPhys.89.041004}.
\newblock URL \url{https://link.aps.org/doi/10.1103/RevModPhys.89.041004}.

\bibitem[Wilde(2013)]{wilde2011classical}
M.~M. Wilde.
\newblock \emph{Quantum information theory}.
\newblock Cambridge university press, 2013.
\newblock \doi{https://doi.org/10.1017/9781316809976}.

\end{thebibliography}
\end{document}